\documentclass[times, review, 10pt]{elsarticle}

\usepackage{amssymb}
\usepackage{amsmath}
\usepackage{amsfonts}
\usepackage{algorithmic}
\usepackage{algorithm}
\usepackage{array}
\usepackage{textcomp}
\usepackage{stfloats}
\usepackage{url}
\usepackage{verbatim}
\usepackage{graphicx}
\usepackage{multirow}
\usepackage{adjustbox}
\usepackage{booktabs}
\usepackage{pifont}
\usepackage{hyperref}
\usepackage{subcaption}
\usepackage{microtype}
\journal{Pattern Recognition}

\begin{document}

\begin{frontmatter}



\title{Test-time Adaptive Hierarchical Co-enhanced Denoising Network for Reliable Multimodal Classification}


\author[inst1,inst2]{Shu~Shen}
\author[inst1,inst2]{C. L.~Philip~Chen}
\author[inst1,inst2]{Tong~Zhang\corref{cor1}}

\cortext[cor1]{Tong Zhang is with the Guangdong Provincial Key Laboratory of Computational AI Models and Cognitive Intelligence, the School of Computer Science and Engineering, South China University of Technology, Guangzhou 510006, China, and is with the Pazhou Lab, Guangzhou 510335, China, and is with Engineering Research Center of the Ministry of Education on Health Intelligent Perception and Paralleled Digital-Human, Guangzhou, China. (e-mail: tony@scut.edu.cn). This work was funded in part by the National Natural Science Foundation of China grant under number 62536004, 62222603, in part by the Key-Area Research and Development Program of Guangdong Province under number 2023B0303030001, in part by the Program for Guangdong Introducing Innovative and Entrepreneurial Teams (2019ZT08X214), and in part by the Science and Technology Program of Guangzhou under number 2024A04J6310, and in part by the Fundamental Research Funds for the Central Universities 2025ZYGXZR021.}


\affiliation[inst1]{organization={The Guangdong Provincial Key Laboratory of Computational Intelligence and Cyberspace Information, the School of Computer Science and Engineering, South China University of Technology},
            city={Guangzhou},
            postcode={510006}, 
            state={Guangdong},
            country={China}}
\affiliation[inst2]{organization={The Pazhou Laboratory},
            city={Guangzhou},
            postcode={510335}, 
            state={Guangdong},
            country={China}}

\begin{abstract}
Reliable learning of multimodal data (e.g., multi-omics) is a widely concerning issue, especially in safety-critical applications such as medical diagnosis. However, low-quality data induced by multimodal noise poses a major challenge in this domain, causing existing methods to suffer from two key limitations. First, they struggle to handle heterogeneous data noise, hindering robust multimodal representation learning. Second, they exhibit limited adaptability and generalization when encountering previously unseen noise. To address these issues, we propose Test-time Adaptive Hierarchical Co-enhanced Denoising Network (TAHCD). On one hand, TAHCD introduces the Adaptive Stable Subspace Alignment and Sample-Adaptive Confidence Alignment to reliably remove heterogeneous noise. They account for noise at both global and instance levels and enable jointly removal of modality-specific and cross-modality noise, achieving robust learning. On the other hand, TAHCD introduces Test-Time Cooperative Enhancement, which adaptively updates the model in response to input noise in a label-free manner, thus improving generalization. This is achieved by collaboratively enhancing the joint removal process of modality-specific and cross-modality noise across global and instance levels according to sample noise. Experiments on multiple benchmarks demonstrate that the proposed method achieves superior classification performance, robustness, and generalization compared with state-of-the-art reliable multimodal learning approaches.
\end{abstract}




\end{frontmatter}


\section{Introduction}
\label{sec:intro}

With the rapid advancement of sensing and data processing technologies, large volumes of multimodal data have become available, spurring interest in multimodal learning and substantially improving performance across many applications \cite{baltruvsaitis2018multimodal,liang2022foundations,LAN2026113025}, such as medical diagnosis \cite{wang2021mogonet,zheng2024global,CUI2026111991} and cross-modality retrieval \cite{li2025multi,ke2025cross,si2025unified}. Data from different modalities provide complementary information that enables more comprehensive and discriminative representations for diverse tasks. However, they also inevitably introduce more complex noise, typically arising from sensor failure, environmental interference, and data collection errors. In real-world scenarios, multimodal noise exhibits two key characteristics: \textbf{\textit{(1) Heterogeneity}}. Previous work \cite{zhang2024multimodal} identifies two main types of multimodal noise: \textbf{modality-specific noise} and \textbf{cross-modality noise}. Modality-specific noise arises independently in each modality from sensor errors or environmental disturbances, while cross-modality noise stems from low-quality data due to weakly aligned or misaligned multimodal samples \cite{changpinyo2021conceptual}. \textbf{\textit{(2) Unpredictability}}. Noise often evolves over time as new samples are incrementally acquired, potentially introducing previously unseen noise. For example, unexpected device failures can introduce unfamiliar noise in newly collected patient data, image–text–mismatched posts continuously emerge in online data streams, and sensors in autonomous driving systems are exposed to varying weather and urban conditions.

\begin{figure}[t]
\centering
\includegraphics[width=0.5\linewidth]{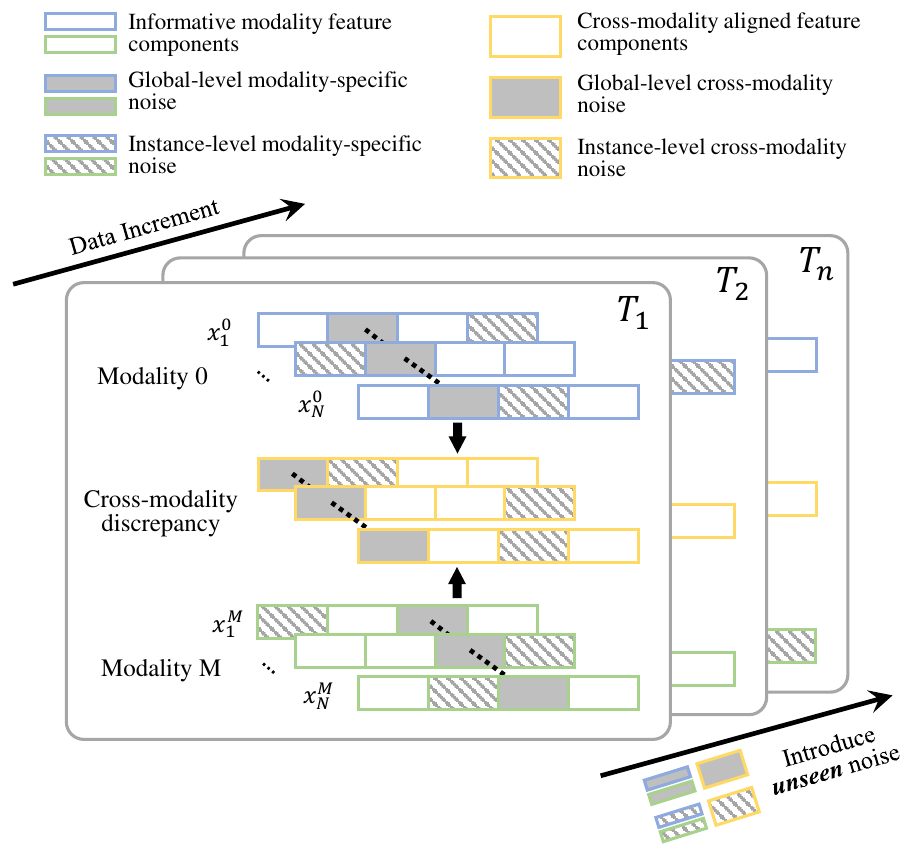} 
\caption{Illustration of complex multimodal noise in real-world scenario. \textbf{(i) Heterogeneity.} Multimodal noise generally falls into two types \cite{zhang2024multimodal}: modality-specific and cross-modality noise, which may occur at the global level across all samples or at the instance level in individual samples. \textbf{(ii) Unpredictability.} With incremental data acquisition, the noise evolves over time and may include previously unseen patterns.}
\label{fig:head}
\end{figure}

In the presence of complex multimodal noise, conventional multimodal methods struggle to produce reliable results, making them unsuitable for safety-critical applications such as disease diagnosis \cite{han2022multimodal,zou2023dpnet,zhou2023calm}. Accordingly, many methods improve the reliability of multimodal learning by removing noise and uninformative feature in data. Among them, \cite{han2022trusted,han2022multimodal,zhou2023calm,zou2023dpnet,zhang2023provable,cao2024predictive} address modality-specific noise via confidence-based feature learning and fusion. However, they generally assume well-aligned inputs and overlook the presence of cross-modality noise. Conversely, \cite{huang2023nlip,nakada2023understanding,zeng2023semantic,xie2025seeking} eliminate cross-modality noise via aligning and rectifying across modalities, but modality-specific noise can easily lead them to learn erroneous and unreliable alignments. Moreover, existing methods typically learn on limited training data, leaving them unaware of unseen noise. Therefore, these methods generally face the following limitations. \textbf{\textit{(1) Insufficient robustness to heterogeneous noise:}} They struggle to jointly handle modality-specific and cross-modality noise. \textbf{\textit{(2) Limited generalization:}} they adapt and generalize poorly to unseen noise.

To address these limitations, we propose Test-time Adaptive Hierarchical Co-enhanced Denoising Network (TAHCD). TAHCD first removes heterogeneous noise by considering it at both global and instance levels. As shown in Figure \ref{fig:head}, data often contain global-level noise with consistent patterns across samples, such as persistent sensor faults or repeated sampling under identical conditions, and instance-level noise affecting individual samples, e.g., sudden sensor failures or human errors. Accordingly, TAHCD proposes Adaptive Stable Subspace Alignment (ASSA) and Sample-Adaptive Confidence Alignment (SACA) to jointly eliminate modality-specific and cross-modality noise at the global and instance levels, respectively, achieving robust representation learning. Moreover, TAHCD incorporates Test-Time Cooperative Enhancement (TTCE) to collaboratively refine global- and instance-level denoising based on input noise in a label-free manner, enabling adaptation to previously unseen noise and substantially improving generalization.

Specifically, ASSA first computes each modality's feature covariance and performs singular value decomposition. A learnable mask conditioned on the singular values then selects eigenvectors to construct stable feature subspaces, guided by inter-class orthogonality and \textbf{subspace projection alignment} constraints. These constraints encourage the mask to discard directions carrying modality-specific and cross-modality noise across all samples. By aligning modality projections in the constructed stable subspaces, the \textbf{subspace projection alignment} in ASSA improves the reliability of cross-modality noise removal by preventing the model from erroneously aligning uninformative semantic content induced by modality-specific noise. Subsequently, SACA uses the globally denoised features from ASSA to estimate priors that guide the experts in removing heterogeneous noise at the instance level via a \textbf{confidence-aware asymmetric slack alignment}. By reweighting gradients based on modality confidence, this strategy encourages low-confidence modalities to align towards high-confidence ones, enabling reliable cross-modality noise removal under instance-level modality-specific noise. It also adopts a slack alignment scheme that constrains cross-modality discrepancies within a reasonable range learned from the prior, reflecting inherent differences in informative content across modalities. Compared with existing methods that strictly enforce cross-modality consistency, this prevents over-suppressing modality-specific complementary information while removing noise, resulting in more robust multimodal representations.

Building upon the above framework for heterogeneous noise removal, we further propose TTCE that adapts the denoising framework to unseen noise in a label-free manner. Since the instance level can capture noise variations more sensitively than the global level, it can also perceive unseen noise more accurately. Based on this consideration, TTCE first incorporates the instance-level modality-specific and cross-modality noise learned by SACA into the reconstruction of globally denoised modality features produced by ASSA. This enables global-level modality representation learning in ASSA to explicitly leverage more accurate and finer-grained unseen-noise information captured at the instance level, thereby enhancing global-level unseen-noise removal. The enhanced global-denoised features in ASSA can provide more reliable priors, which in turn enhance instance-level unseen-noise learning and removal in SACA. Through multiple iterations of the above co-enhancement process between ASSA and SACA, the model achieves progressive update and adaptation to unseen noise.

The contributions of this paper can be summarized as:
\begin{itemize}
    \item We propose TAHCD, which jointly removes modality-specific and cross-modality noise across global and instance levels via ASSA and SACA while adapting to previously unseen noise. It achieves SOTA results across multiple benchmarks under diverse noise conditions.
    \item ASSA's subspace projection alignment and SACA's confidence-aware asymmetric slack alignment mitigate modality-specific noise interference and preserve complementary modality feature during cross-modality noise removal, facilitating joint removal of both noise types.
    \item The proposed TTCE in TAHCD cooperatively enhances ASSA and SACA driven by input sample noise in a label-free manner. This enables adaptive improvement of the denoising process at test time for previously unseen noise, thereby enhancing generalization ability.
\end{itemize}

\section{Related Works}
This section briefly reviews related works on multimodal learning and reliable learning from noisy, low-quality multimodal data.
\subsection{Multimodal Learning}
\label{sec:relative-mml}

The advancement of sensor technology and the development of data transmission methods have driven the rapid growth of multimodal learning in a wide range of tasks and applications \cite{baltruvsaitis2018multimodal,zhu2024vision+}. For instance, \cite{lee2021variational,wang2021mogonet,zheng2024global,CUI2026111991} have achieved precise pathological classification by integrating multiple omics data. \cite{li2025multi,ke2025cross,si2025unified} have developed cross-modality retrieval methods based on modalities such as images and text. \cite{9906429,zhang2025prompts} have constructed multimodal networks based on images and text, improving performance on computer vision tasks such as classification and generation. However, in real-world scenarios, the prevalence of low-quality data caused by noise substantially undermines the reliability of multimodal methods \cite{zhang2024multimodal,zhang2023provable}, with particularly severe consequences in safety-critical applications such as medical diagnosis. Therefore, this work focuses on low-quality data and develops multimodal learning method that is robust and generalizable to multimodal noise.

\subsection{Reliable Learning from Low-Quality Multimodal Data}
In real-world scenarios, multimodal learning often suffers from low-quality data, which undermines reliability. A major source of low-quality data is multimodal noise, typically arising from factors such as sensor failures, environmental influences, or human errors. Existing studies \cite{zhang2024multimodal} categorize multimodal noise into two types, namely modality-specific noise and cross-modality noise, and have developed various methods to address these noise types for reliable multimodal learning. 

Many studies mitigate modality-specific noise through quality-based representation learning and fusion methods. Geng et al. \cite{geng2021uncertainty} have proposed the DUA-Nets, which achieved uncertainty-based multimodal representation learning through reconstruction. Han et al. \cite{han2022multimodal} have modeled informativeness at the feature and modality levels, achieving trustworthy multimodal feature fusion. Zhang et al. \cite{zhang2023provable} have achieved more robust multimodal fusion by dynamically assigning weights to each modality based on uncertainty estimation. Zheng
et al. \cite{zheng2023multi} have achieved trustworthy multimodal classification via integrating feature and label-level confidence. Zhou et al. \cite{zhou2023calm} have introduced a trustworthy multi-view classification framework via confidence-aware fusion. Cao et al. \cite{cao2024predictive} have proposed adjusting modality weights based on their losses to achieve robust fusion. However, these methods assume that the modalities are well-aligned and therefore tend to yield unreliable results in the presence of cross-modality noise. Conversely, many studies delve into cross-modality noise removal by modeling cross-modality alignment and correspondence. \cite{radenovic2023filtering,gadre2024datacomp} have proposed rule-based methods to remove misaligned or weakly aligned data. \cite{huang2021learning,dang2024noisy,xie2025seeking,wang2025noisy} have designed models to filter or correct misaligned samples. \cite{huang2023nlip,zeng2023semantic,dang2025disentangled} have proposed regularization methods to mitigate the impact of misaligned samples. However, these methods are easily misled by modality-specific noise, leading to erroneous and unreliable alignment learning. Moreover, existing methods typically learn and update on limited training data, hindering adaptation and generalization to unseen and varying noise conditions. Therefore, we propose a multimodal learning method that jointly handles modality-specific and cross-modality noise and enables test-time adaptive updates, thus enhancing robustness to heterogeneous noise and generalization to unseen noise.

\section{Proposed Approach} \label{sec:prop_frame}




In this section, we provide a detailed introduction of the proposed Test-time Adaptive Hierarchical Co-enhanced Denoising Network (TAHCD). Let $\mathcal{D}=\{(x_i, y_i)\}_{i=1}^N$ denote a multimodal dataset with $N$ samples, where each sample contains $M$ modalities and is assigned one of $C$ class labels. The objective of multimodal classification is to train a neural network that maps each input $x_i=\{x_i^{m}\in\mathbb{R}^{d^{m}}\}_{m=1}^{M}$ to its corresponding label $y_i\in\mathbb{R}^{C}$, with $d^{m}$ representing the feature dimension of the $m$-th modality. However, dataset \(\mathcal{D}\) typically contains complex multimodal noise, which undermines the reliability of multimodal classification. To this end, we propose TAHCD, which jointly removes modality-specific and cross-modality noise at global and instance levels and adaptively updates to input noise in a label-free manner, enabling robust and generalizable multimodal classification.

\begin{figure*}[t]
\centering
\includegraphics[width=\linewidth]{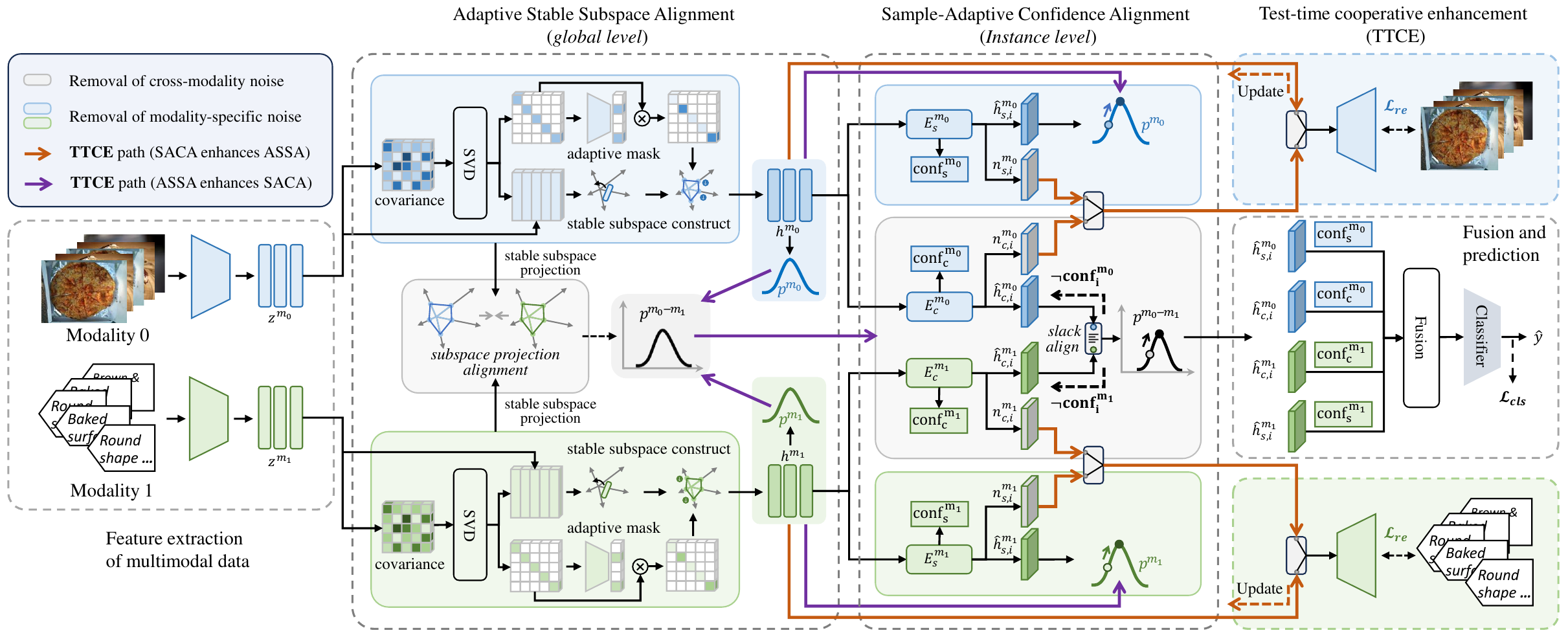} 
\caption{The framework diagram of the Test-time Adaptive Hierarchical Co-enhanced Denoising Network (TAHCD), best viewed in color.
TAHCD consists of three key components: (1) Adaptive Stable Subspace Alignment (ASSA): jointly removes modality-specific and cross-modality noise at the global level. (2) Sample-Adaptive Confidence Alignment (SACA): jointly removes modality-specific and cross-modality noise at the instance level. (3) Test-Time Cooperative Enhancement (TTCE): adaptively enhances the model in response to test-time noise. Without loss of generality, the diagram illustrates a scenario with two modalities, where the blue and green colors represent two distinct modalities.}
\label{fig:framework_overall}
\end{figure*}


\subsection{Overview of TAHCD}
As illustrated in Figure \ref{fig:framework_overall}, TAHCD first employs modality-specific feature encoders to extract representations from each modality. It then introduces Adaptive Stable Subspace Alignment (ASSA) to remove modality-specific and cross-modality noise exhibiting global distributional characteristics. In addition, Sample-Adaptive Confidence Alignment (SACA) is proposed to eliminate modality-specific and cross-modality noise at the instance level. Building upon these components, we further design Test-Time Cooperative Enhancement (TTCE) mechanism. It achieves cooperative enhancement between global- and instance-level denoising, driven by sample noise in a label-free manner, improving adaptability and generalization to unseen noise.

\subsection{Adaptive Stable Subspace Alignment} \label{sec:assa}
To eliminate modality-specific noise and cross-modality noise with global characteristics across all samples in multimodal features, the Adaptive Stable Subspace Alignment (ASSA) strategy is proposed. ASSA adaptively selects highly informative principal axes in the feature space to construct a stable feature subspace. Subsequently, inter-class orthogonality constraints within each modality are imposed to ensure that the learned subspaces effectively suppress modality-specific noise. Furthermore, we introduce a subspace projection alignment constraint to prevent erroneous alignment of uninformative semantic content under modality-specific noise and enhance cross-modality noise removal.

\subsubsection{Adaptive Stable Subspace Construction}

To facilitate global-level multimodal noise removal, we adaptively eliminate statistically unstable and uninformative directions in the feature distribution of each modality and suppress noise-dominated variations across samples. Specifically, for the \(m\)-th modality, the input \(x^{m}=\{x_1^m,\dots,x_N^m\}\) is first encoded into the latent space \(z^{m}=\{z_1^m,\dots,z_N^m\}\) by the corresponding modality encoder \(\phi_x^m\): $z^m=\phi_x^m(x^m)$. Then, the covariance matrix \(\Sigma^{m}_z\) among features is computed from \(z^{m}\), which can be formulated as:
\begin{align}
    \Sigma^m_z&=\mathbb{E}\left[(z^m-\mu^m_z)(z^m-\mu^m_z)^{\top}\right]\in \mathbb{R}^{d^m\times d^m},
    \label{eq:Sigma}
\end{align}
where
\begin{align}
        \mu_z^m &= \frac{1}{N}\sum_{i=1}^N z_i^m \in \mathbb{R}^{d_m}.
        \label{eq:mu}
\end{align}
Subsequently, we perform singular value decomposition on the covariance matrix \(\Sigma^{m}_z\), which can be expressed as:
\begin{align}
    \Sigma^m_z=U^m_z\Lambda^m_z(U^m_z)^{\top}.
\end{align}
\(U^m_z = [u^m_1, \ldots, u^m_{d^m}]\) is the matrix of orthogonal eigenvectors, and \(\Lambda^m_z = \mathrm{diag}(\lambda_1^m, \ldots, \lambda_{d^m}^m)\) contains the non-negative singular values. Each vector \(u_i^m(i\in[1,d_m])\) in \(U_z^m\) represents a principal axis in the feature space, while the corresponding singular value \(\lambda_i^m\) denotes the variance along that axis and is closely related to its informativeness. Therefore, we propose to learn a mask \(w_{\lambda}^m\) from the singular values \(\lambda^m = [\lambda_1^m, \ldots, \lambda_{d^m}^m]\) to adaptively select informative principal axes while suppressing noisy and redundant ones, thereby achieving global-level noise removal on $z^m$. This process can be formulated as:
\begin{align}
    w^m_{\lambda}&=\sigma(\phi^m_{\lambda}(\lambda^m))\in \mathbb{R}^{d^m},\\
    h^m&=z^mU^m_z\mathrm{diag}(w^m_{\lambda})(U^m_z)^{\top}. \label{eq:adaptive-filter}
\end{align}
$\phi_{\lambda}^m(\cdot)$ is a learnable encoder, \(\sigma(\cdot)\) denotes the sigmoid function. \(h^m=\{h_1^m,\dots,h_N^m\}\) denote the globally denoised representation obtained from the latent feature \(z^m\) of modality $m$. As shown in Eq. (\ref{eq:adaptive-filter}), the feature vector \(z^m\) is first projected onto the covariance principal axes using the orthogonal basis \(U_z^m\). It is then filtered by the learned mask \(w^m_{\lambda}\) and finally mapped back to the original dimension via \((U^m_z)^\top\).

\subsubsection{Constraints}
To ensure that the adaptive stable subspace construction sufficiently eliminates modality-specific and cross-modality noise, we introduce  inter-class orthogonality constraint \(\mathcal{L}_{\mathrm{o}}\) and subspace projection alignment constraint \(\mathcal{L}_{\mathrm{a}}\). The inter-class orthogonality constraint \(\mathcal{L}_{\mathrm{o}}\) eliminates modality-specific noise by removing the shared spurious patterns it induces across classes within the same modality, while preserving discriminative features of different classes, which can be formulated as:
\begin{align}
    \mathcal{L}_{\mathrm{o}}=\frac{1}{C(C-1)}\sum_{m=1}^M\sum_{c=1}^C\sum_{c^\prime\neq c}^C\|(\mu^{m,c})^\top\cdot \mu^{m,c^\prime}-I\|_{\mathrm{F}}^2.
\end{align}
$\mu^{m,c}=\frac{1}{|x^{m,c}|}\sum_{i=1}^{|x^{m,c}|}x_i^{m,c}$, where $x^{m,c}=\{x_i^{m,c}\}_{i=1}^{|x^{m,c}|}$ denotes the set of modality-\(m\) features for all samples belonging to class \(c\), and \(|\cdot|\) represents the cardinality of the set. \(I \in \mathbb{R}^{C \times C}\) denotes the identity matrix. 

Additionally, the subspace projection alignment constraint \(\mathcal{L}_{\mathrm{a}}\) removes cross-modality noise by aligning the projections of representations from different modalities within their respective stable subspaces, as shown in Eq. (\ref{eq:align-loss}). By projecting modality features into their respective stable subspaces before alignment, it mitigates the impact of noise in each modality on alignment learning, ensuring stable and reliable cross-modality correspondences.
\begin{align}
    \mathcal{L}_{a}=\sum_{m=1}^M\sum_{m^{\prime}\neq m}^M \|z^mU^m_z\mathrm{diag}(w^m_{\lambda})-z^{m^{\prime}}U^{m^{\prime}}_z\mathrm{diag}(w^{m^{\prime}}_{\lambda})\|_{\mathrm{F}}^2.
    \label{eq:align-loss}
\end{align}
Based on the above constraints,  we minimize the total loss $\mathcal{L}_{\mathrm{assa}}$ defined in Eq. (\ref{eq:l-assa}) to remove global-level noise in ASSA.
\begin{align}
    \mathcal{L}_{\mathrm{assa}}=\mathcal{L}_{\mathrm{o}}+\mathcal{L}_{\mathrm{a}}. \label{eq:l-assa}
\end{align}


\subsection{Sample-Adaptive Confidence Alignment}\label{sec:smne}
To provide reliable supervision signals for instance-level noise removal, we first estimate modality-wise feature distributions and cross-modality discrepancy distributions using the globally denoised representations obtained from ASSA. These distributions are then serve as priors to guide sample-adaptive noise experts with confidence-aware asymmetric slack alignment for instance-level noise removal. 

\subsubsection{Prior Estimation}
Based on the representations \(\{h^m\}_{m=1}^M\) of different modalities obtained from Eq. (\ref{eq:adaptive-filter}) in ASSA, we estimate the modality-wise feature distributions $\{h^m\sim p^m_h=\mathcal{N}(\mu^m_h,\Sigma^m_h)\}_{m=1}^m$ and cross-modality discrepancy distributions $\{\{(h^m-h^{m^\prime})\sim p^{m-m^\prime}_h=\mathcal{N}(\mu^{m-m^\prime}_h,\Sigma^{m-m^\prime}_h)\}_{m^\prime\neq m}^M\}_{m=1}^M$. Specifically, for the modality-wise feature distribution $p^m_h$, we use maximum likelihood estimation to calculate $\mu^m_h$ and $\Sigma^m_h$ from $h^m$, which can be obtained by replacing \( z^m \) in Eq. (\ref{eq:mu}) and Eq. (\ref{eq:Sigma}) with \( h^m \), respectively. Subsequently, we compute the distribution of \((h^m - h^{m^\prime})\) between any two distinct modalities \(m\) and \(m^\prime\) as the cross-modality discrepancy distribution $p^{m-m^\prime}_h$, whose mean \(\mu^{m-m^\prime}_h\) and covariance \(\Sigma^{m-m^\prime}_h\) can be obtained by:
\begin{align}
    \mu^{m-m^{\prime}}_h&=\mu^m_h-\mu^{m^\prime}_h\notag \\
    \Sigma^{m-m^{\prime}}_h&=\Sigma^m_h+\Sigma^{m^\prime}_h 
    \label{eq:modality-dis-distri-cal}
\end{align}



\subsubsection{Sample-Adaptive Noise Expert}
Modality-specific noise experts and cross-modality noise experts are then introduced to perform noise removal on each sample under the guidance of the estimated priors. Specifically, we introduce \(M\) modality-specific noise experts \({E_s^1, \ldots, E_s^M}\) and \(M\) cross-modality noise experts \({E_c^1, \ldots, E_c^M}\), each corresponding to one modality. For a multimodal sample \(x_i=\{x_i^m\}_{m=1}^M\) with globally denoised representation \(h_i=\{h_i^m\}_{m=1}^M\), each modality representation \(h_i^m\) is fed into \(E_s^m\) and \(E_c^m\) to learn masks \(w_{s,i}^m\) and \(w_{c,i}^m\) that capture feature-wise informativeness relevant to modality-specific and cross-modality noise, respectively. These masks are then used to remove the modality-specific and cross-modality noise components \(n_{s,i}^m, n_{c,i}^m\) from $h_i^m$ and yield the corresponding denoised features \(\hat{h}_{s,i}^m, \hat{h}_{c,i}^m\). The calculation in $E_s^m$ can be formulated as:
\begin{align}
    &w_{s,i}^m=\sigma(\phi_{s}^m(h_i^m))\in \mathbb{R}^{d^m},\notag\\
    &\hat{h}_{s,i}^m=h_i^m\odot w_{s,i}^m,\quad n_{s,i}^m=h_i^m\odot (1-w_{s,i}^m). \label{eq:exp-msn-remove}
\end{align}
The calculation in $E_c^m$ can be formulated as:
\begin{align}
    &w_{c,i}^m=\sigma(\phi_{c}^m(h_i^m))\in \mathbb{R}^{d^m},\notag\\
    &\hat{h}_{c,i}^m=h_i^m\odot w_{c,i}^m,\quad n_{c,i}^m=h_i^m\odot (1-w_{c,i}^m). \label{eq:exp-cmn-remove}
\end{align}
\(\phi_s^m\) and \(\phi_c^m\) denote the learnable networks in \(E_s^m\) and \(E_c^m\), respectively, and \(\sigma\) represents the sigmoid function. The operator \(\odot\) denotes element-wise multiplication. 

\subsubsection{Confidence-Aware Asymmetric Slack Alignment}
We minimize the negative log-likelihood \(\mathcal{L}_{\mathrm{nll}}^{s}\) for \(\hat{h}_s^{m}=\{\hat{h}_{s,i}^m\}_{i=1}^N\) under \(p_h^{m}\), and \(\mathcal{L}_{\mathrm{nll}}^{c}\) for the cross-modality discrepancy \(\hat{h}_c^{m-m^\prime}=\{h_{c,i}^{m} - h_{c,i}^{m^\prime}\}_{i=1}^N\) under the cross-modality discrepancy distribution \(p_h^{m-m^\prime}\), to respectively ensure effective removal of modality-specific and cross-modality noise in each sample. The overall objective $\mathcal{L}_{\mathrm{saca}}$ is formulated as:
\begin{align}
    &\mathcal{L}_{\mathrm{saca}}=\mathcal{L}_{\mathrm{nll}}^{s}+\mathcal{L}_{\mathrm{nll}}^{c}=\left[-\sum_{m=1}^M\log p^m_h(\hat{h}_s^m)\right]+\left[-\sum_{m=1}^M\sum_{m^\prime> m}^M\log p^{m-m^\prime}_h(\hat{h}_c^m-\hat{h}_c^{m^\prime})\right].
\label{eq:l-sam}
\end{align}
It is worth noting that, in \(\mathcal{L}_{\mathrm{nll}}^{c}\), unlike existing methods that remove cross-modality noise by maximizing similarity across modalities, we instead constrain the feature discrepancies $(\hat{h}_c^m-\hat{h}_c^{m^\prime})$ across modalities within a reasonable range. This range accounts for the inherent informative information differences across modalities, which is learned by the cross-modality discrepancy distribution $p_h^{m-m^\prime}$. We refer to this strategy as \textbf{\textit{slack alignment}}, which enables the removal of cross-modality noise while preserving more modality-specific complementary information, thereby facilitating more reliable multimodal representation learning. 
Moreover, considering confidence differences across modalities under modality-specific noise, we implement the slack alignment in an \textbf{\textit{asymmetric manner directed by modality confidence}}. This encourages low-confidence modality features with more modality-specific noise to align toward high-confidence ones, imposing stronger rectification on them, instead of rectifying uniformly across all modalities. Specifically, the update of the parameters in cross-modality noise expert \(E_c^m\) under the guidance of modality confidence can be expressed as:
\begin{align}
    \theta_c^m\leftarrow \theta_c^m-\eta \cdot \frac{1}{|B|}\sum_i^{|B|}\neg conf^m_i\cdot \nabla_{\theta_c^m}\mathcal{L}_{\mathrm{nll}}^c(\hat{h}_c^m;\theta_c^m). \label{eq:aym-conf}
\end{align}
\(\theta_c^m\) denotes the parameters of the cross-modality noise expert \(E_c^m\) corresponding to modality $m$. \(\eta\) is the learning rate, $B$ is a random mini-batch, and \(\nabla_{\theta_c^m}\mathcal{L}_{\mathrm{nll}}^c\) is the gradient of \(\theta_c^m\) with respect to \(\mathcal{L}_{\mathrm{nll}}^c\). \(\neg conf^m_i\) is a factor negatively correlated with the confidence \(c_i^m\) of modality-$m$ feature \(h^m_i\) and is dynamically computed during training as follows:
\begin{align}
    \neg conf_i^m=\frac{\exp(1-\mathrm{tanh}(c_i^m))}{\sum_{m^\prime=1}^{M}\exp(1-\mathrm{tanh}(c_i^{m^\prime}))},
\end{align}
where $c_i^m=p^m(h_i^m)$ is estimated as the likelihood of $h_i^m$ under under the modality-$m$ feature prior $p^m$. In this way, the update gradients of cross-modality noise experts are amplified for low-confidence modality and reduced for high-confidence ones. This encourages the model to focus on mining and applying stronger alignment rectification to lower-confidence modalities with more pronounced modality-specific noise. Consequently, the model achieves more reliable cross-modality noise mitigation under modality-specific noise, enhancing the joint removal of both noise types.

\subsection{Test-Time Cooperative Enhancement} \label{sec:ttce}
ASSA and SACA establish a framework for jointly removing modality-specific and cross-modality noise at both global and individual levels to achieve robust learning. Building on this framework, we introduce Test-Time Cooperative Enhancement (TTCE), which performs label-free adaptive updates guided by sample noise to improve generalization to unseen noise through cooperative enhancement of ASSA and SACA.

Considering that instance-level representations are more sensitive to noise variations than global distributions, noise information learned at the instance level can capture richer cues about unseen noise than that at the global level. Therefore, TTCE proposes to first leverage instance-level noise learned by SACA to enhance global-level denoising of ASSA under unseen noise. Specifically, we incorporate instance-level modality-specific and cross-modality noise components $n_s^m=\{n_{s,i}^m\}_{i=1}^N$ and $n_c^m=\{n_{c,i}^m\}_{i=1}^N$ from SACA into the globally denoised modality features $h^m$ from ASSA and define a reconstruction loss:
\begin{align}
    \mathcal{L}_{\mathrm{re}}=\frac{1}{M}\sum_{m=1}^M{\|\Psi^m(h^m+n_s^m+n_c^m)-x^m\|_{\mathrm{F}}^2}. \label{eq:l-re}
\end{align}
\(\Psi^m(\cdot)\) denotes the decoder corresponding to modality \(m\), and \(\|\cdot\|_{\mathrm{F}}^2\) denotes the Frobenius norm. Minimizing $\mathcal{L}_{\mathrm{re}}$ enables ASSA to explicitly account for more accurate and finer-grained unseen-noise information captured at the instance level, yielding more reliable globally denoised modality representations \(h^{m\uparrow}\) to unseen noise.
The process of enhancing $h^m$ to $h^{m\uparrow}$ via minimizing $\mathcal{L}_{\mathrm{re}}$ can be equivalently expressed as:
\begin{align}
    h^{m\uparrow}&=h^{m\uparrow}-\eta \frac{\partial \mathcal{L}_{\mathrm{re}}}{\partial h^{m\uparrow}}=h^{m\uparrow}-2\eta J_{\Psi^m}(I^m)^{\top}(\Psi^m(I^m)-x^m),
    \label{eq:sam-updata-glb}
\end{align}
where $h^{m\uparrow}$ is initialized as $h^m$ at the beginning of the update process. $I^m=h^m+n_s^m+n_c^m$. $J_{\Psi^m}$ represents the Jacobian matrix of the pre-trained decoder $\Psi^m$ evaluated at the input. $\eta$ is the updating rate. 

Moreover, the enhanced modality features \(h^{m\uparrow}\) in ASSA can help remove unseen noise from the priors, which in turn facilitating instance-level denoising in SACA. Specifically, for the modality-wise distribution $p^m\sim\mathcal{N}(\mu^m,\Sigma^m)$, we compute \(\Delta \mu^{m\uparrow}\) and \(\Delta \Sigma^{m\uparrow}\) by substituting \(z^m\) with \(h^{m\uparrow}\) in Eqs. (\ref{eq:mu}) and (\ref{eq:Sigma}), respectively, and use them to update $\mu^m$ and $\Sigma^m$:
\begin{align}
    \mu^{m\uparrow}&=(1-\alpha)\cdot \mu^m+\alpha\cdot \Delta\mu^{m\uparrow},\notag\\
    \Sigma^{m\uparrow}&=(1-\alpha)\cdot \Sigma^m+\alpha\cdot \Delta\Sigma^{m\uparrow}. \label{eq:update-mu-Sigma}
\end{align}
$\mu^{m\uparrow}$ and $\Sigma^{m\uparrow}$ are the mean and covariance of the updated modality-wise distribution $p^m$, denoted as $p^{m\uparrow}$. $\alpha\in[0,1]$ is a hyperparameter, which will be analyzed in Sec. \ref{sec:pa}. Accordingly, the cross-modality discrepancy distribution $p^{m-m^\prime}(m,m^\prime\in[1,M],m\neq m^\prime)$ is updated to $p^{m-m^\prime\uparrow}\sim\mathcal{N}(\mu^{m-m\prime\uparrow},\Sigma^{m-m^\prime\uparrow})$, with $\mu^{m-m\prime\uparrow}$ and $\Sigma^{m-m^\prime\uparrow}$ derived from $\mu^{m\uparrow}$, $\mu^{m\prime\uparrow}$, $\Sigma^{m\uparrow}$, and $\Sigma^{m\prime\uparrow}$ through Eq. (\ref{eq:modality-dis-distri-cal}). The enhanced priors $p^{m\uparrow}$, $p^{m\prime\uparrow}$, and $p^{m-m\prime\uparrow}$ are then applied in $\mathcal{L}_{\mathrm{saca}}$ defined in Eq. (\ref{eq:l-sam}) to improve the instance-level noise removal of the noise experts against unseen noise, which can be equivalently expressed as:
\begin{align}
    &\hat{h}_s^{m\uparrow}=\hat{h}_s^{m\uparrow}-\eta \frac{\partial\mathcal{L}_{\mathrm{saca}}}{\partial \hat{h}_s^{m\uparrow}}=\hat{h}_s^{m\uparrow}-\eta(\Sigma^{m\uparrow})^{-1}(\hat{h}_s^{m\uparrow}-\mu^{m\uparrow}), \notag\\
    &\hat{h}_c^{m\uparrow}=\hat{h}_c^{m\uparrow}-\eta \frac{\partial\mathcal{L}_{\mathrm{saca}}}{\partial \hat{h}_c^{m\uparrow}}=\hat{h}_c^{m\uparrow}-\eta\sum\nolimits_{m^{\prime}\neq m}^M{(\Sigma^{m-m^{\prime\uparrow}})^{-1}(\hat{h}_c^{m\uparrow}-\hat{h}_c^{m\prime\uparrow}-\mu^{m-m^{\prime\uparrow}})}.
\label{eq:instance-update}
\end{align}

Therefore, through multiple iterations of the above cooperative enhancement between ASSA and SACA, TTCE progressively updates the model to adapt to unseen noise. For clarity, the overall inference procedure of TAHCD with TTCE is summarized in Algorithm \ref{alg:algorithm}, where \(E\) denotes the number of cooperative enhancement iteration between ASSA and SACA and will be analyzed in Sec. \ref{sec:dis-E}.

\begin{algorithm}[t]
\caption{Inference procedure of TAHCD with TTCE}
\label{alg:algorithm}
\textbf{Input}: Multimodal input $X=\{\{x_i^m\}_{i=1}^N\}_{m=1}^M$, trained TAHCD.

\begin{algorithmic}[1] 
\STATE Compute \(h^m\) with ASSA by Eq. (\ref{eq:adaptive-filter}).
\STATE Estimate priors \(p^m, p^{m-m^\prime}\) by Eqs. (\ref{eq:Sigma}), (\ref{eq:mu}), and (\ref{eq:modality-dis-distri-cal}).
\STATE Compute \(\hat{h}_s^m, \hat{h}_c^m, n_s^m, n_c^m\) in SACA by Eqs. (\ref{eq:exp-msn-remove}), (\ref{eq:exp-cmn-remove}).
\FOR{$e=1,...,E$}
\STATE Calculate $\mathcal{L}_{\mathrm{re}}$ using Eq. (\ref{eq:l-re}).
\STATE Enhance ASSA and $h^{m\uparrow}$ using Eq. (\ref{eq:sam-updata-glb}).
\STATE Update priors \(p^{m\uparrow}, p^{m-m^\prime\uparrow}\) using Eq. (\ref{eq:update-mu-Sigma}) and (\ref{eq:modality-dis-distri-cal}).
\STATE Enhance SACA and \(\hat{h}_s^{m\uparrow}, \hat{h}_c^{m\uparrow}\) using Eq. (\ref{eq:instance-update}).
\ENDFOR
\STATE Fuse $\{\hat{h}_s^{m\uparrow}\}_{m=1}^M$, $\{\hat{h}_c^{m\uparrow}\}_{m=1}^M$ using Eq. (\ref{eq:fusion}) for prediction.
\end{algorithmic}
\label{alg:aim}
\end{algorithm}

\subsection{Training and Inference} \label{sec:fp}
Finally, we fuse the multimodal features $\{h_s^{m\uparrow}\}_{m=1}^M,\{h_c^{m\uparrow}\}_{m=1}^M$ produced by sample-adaptive noise experts after TTCE, weighted by their respective confidence scores. Specifically, the confidence score is first estimated as the likelihood of the feature vector under the corresponding modality-wise distribution: $\mathrm{conf}_s^{m\uparrow} = p^{m\uparrow}(\hat{h}_s^{m\uparrow}), \mathrm{conf}_c^{m\uparrow} = p^{m\uparrow}(\hat{h}_c^{m\uparrow})$. We then normalize the confidence scores of all experts to obtain weights $\{\overline{\mathrm{conf}}_s^{m\uparrow}\}_{m=1}^M, \{\overline{\mathrm{conf}}_c^{m\uparrow}\}_{m=1}^M$, which are used to fuse $\{h_s^{m\uparrow}\}_{m=1}^M,\{h_c^{m\uparrow}\}_{m=1}^M$:
\begin{align}
    f^{mm}=\sum_{m=1}^M(\overline{\mathrm{conf}}_s^{m\uparrow}\cdot \hat{h}_s^{m\uparrow}+\overline{\mathrm{conf}}_c^{m\uparrow}\cdot \hat{h}_c^{m\uparrow}). \label{eq:fusion}
\end{align}
The obtained multimodal feature \(f^{mm}\) is fed into a classifier to produce the prediction \(\hat{y}\), and the loss is computed using the cross-entropy with the ground-truth label: \(\mathcal{L}_{\mathrm{cls}} = \mathrm{CE}(y,\hat{y})\), where $\mathrm{CE}(\cdot)$ is the cross-entropy function. TAHCD is trained using the total loss \(\mathcal{L}_{\text{tot}}=\mathcal{L}_{\text{assa}} + \mathcal{L}_{\text{saca}} + \mathcal{L}_{\text{re}} + \mathcal{L}_{\text{cls}}\), where the prediction is obtained from modality features after \(E\) iterations of enhancement in TTCE. \(\mathcal{L}_{\text{assa}}, \mathcal{L}_{\text{saca}}\), and \(\mathcal{L}_{\text{re}}\) are defined in Eqs. (\ref{eq:l-assa}), (\ref{eq:l-sam}), and (\ref{eq:l-re}), respectively. The inference procedure during testing of TAHCD is shown in Algorithm \ref{alg:aim}.

\section{Experiments}
In this section, we conduct extensive experiments on four datasets to evaluate the robustness and generalization of the proposed Test-time Adaptive Hierarchical Co-enhanced Denoising Network (TAHCD) under multimodal noise, comparing it with eight representative reliable multimodal learning methods for low-quality data.

\subsection{Experimental Settings} \label{sec:exp-set}

\subsubsection{Datasets}
Experiments are conducted on four commonly used multimodal datasets in previous works \cite{han2022multimodal,zhang2023provable,zheng2023multi,zhou2023calm,pmlr-v235-cao24c}, including two multi-omics datasets. \textbf {(1) BRCA}: BRCA \cite{lingle9cancer} is a breast invasive carcinoma dataset for PAM50 subtype classification, comprising 875 samples with features from three modalities: mRNA expression (mRNA), DNA methylation (meth), and miRNA expression (miRNA). Samples are categorized into five subtypes: Normal-like, Basal-like, HER2-enriched, Luminal A, and Luminal B. BRCA is available from The Cancer Genome Atlas (TCGA) \footnote{\url{https://www.cancer.gov/aboutnci/organization/ccg/research/structuralgenomics/tcga}}. \textbf{(2) ROSMAP}: ROSMAP \cite{mukherjee2015religious,a2012overview,de2018multi} comprises 351 samples from Alzheimer's patients and normal controls, each with three modalities: mRNA expression (mRNA), DNA methylation (meth), and miRNA expression (miRNA). \textbf{(3) CUB}: The Caltech-UCSD Birds dataset \cite{wah2011caltech} contains 11,788 samples across 200 bird categories, each with two modalities: bird images and corresponding textual descriptions. \textbf{(4) UPMC FOOD101}: The UPMC FOOD101 dataset \cite{wang2015recipe} contains 90,704 samples from 101 food categories, each with an image obtained via Google Image search and a corresponding text description collected from uncontrolled environments, thus inherently containing noise.

\subsubsection{Compared Methods}
To validate the reliability of TAHCD under multimodal noise, we compare it with several representative reliable multimodal learning methods in our experiments, including multimodal dynamics (\textbf{MD}) \cite{han2022multimodal}, multi-level confidence learning (\textbf{MLCLNet}) \cite{zheng2023multi}, quality-aware multimodal fusion (\textbf{QMF}) \cite{zhang2023provable}, predictive dynamic fusion (\textbf{PDF}) \cite{cao2024predictive}, Semantic Invariance Learning (\textbf{SMILE}) \cite{zeng2023semantic}, Seeking Proxy Point via Stable Feature Space (\textbf{SPS}) \cite{xie2025seeking}, Asymmetric Similarity Learning (\textbf{ASL}) \cite{wang2025noisy}, and Disentangled Noisy Correspondence Learning (\textbf{DisNCL}) \cite{dang2025disentangled}. The first four methods focus on learning high-confidence modality representations or fusing different modalities based on their qualities, and are therefore regarded as representative approaches for modality-specific noise removal. The remaining four methods emphasize learning cross-modality consistency and correspondence, and are thus treated as representative approaches for cross-modality noise removal.

\subsubsection{Evaluation Metrics}
\paragraph{BRCA \& CUB \& UMPC FOOD101} BRCA, CUB and UMPC FOOD101 provide multi-class classification task. Three metrics, including accuracy (ACC), average F1 score weighted by support (WeightedF1), and macro-averaged F1 score (MacroF1), are employed to evaluate the performance of different methods. \paragraph{ROSMAP} ROSMAP provides a binary classification task. The accuracy (ACC), F1 score (F1), and area under the receiver operating characteristic curve (AUC) of different methods are reported by experiment.

\subsubsection{Implementation Details}
\paragraph{Training Details} We implement the proposed method and other compared methods on the PyTorch 1.12.0 and cuda 11.6 platform, running on Ubuntu 20.04.2 LTS, utilizing one GPU (NVIDIA RTX A6000 with 48 GB of memory) and CPU of AMD EPYC 75F3. The Adam optimizer with learning rate decay is employed to train the model. The initial learning rate of the Adam optimizer is set to 1e-4, the weight decay is set to 1e-4, and the multiplicative factor of the learning rate decay is set to 0.2. All the quantitative results of the proposed TAHCD are the average of five random seeds. 

\paragraph{Model Implementation Details} In TAHCD, the modality encoder \(\phi_x^m\) is a fully connected network with input dimension equal to the modality dimension of the dataset and output dimension 256. The encoders \(\phi_{\lambda}^m\), \(\phi_s^m\), and \(\phi_c^m\) are fully connected networks with input and output dimensions of 256 and a hidden layer of 512. The decoder \(\psi^m\) is a fully connected network with input dimension 256 and output dimension equal to the modality dimension of the dataset.

\paragraph{Experimental Details} Two types of noise are involved in the experiments: modality-specific and cross-modality noise. Modality-specific noise is implemented following \cite{geng2021uncertainty,zhou2023calm,cao2024predictive}, with severity level \(\epsilon\). Cross-modality noise follows \cite{zeng2023semantic,dang2025disentangled}, where unaligned samples are introduced by shuffling the features of each modality for a random subset of samples with a proportion of $\eta$.

\subsection{Comparison with State-of-the-Art Methods} \label{sec:comp-sota}
To evaluate the robustness and generalization of TAHCD under multimodal noise, we conduct experiments on datasets corrupted by modality-specific and cross-modality noise \cite{han2022multimodal,cao2024predictive,zeng2023semantic,dang2025disentangled}, with \(\epsilon\) and \(\eta\) denoting their respective severity levels. 
\subsubsection{Robustness} To evaluate the improved robustness of TAHCD against multimodal noise, we compare the classification performance of different methods on train and test sets that are both corrupted by the same noise, as reported in Table \ref{tab:sota}. MD, MLCLNet, QMF, and PDF learn reliable modality representations by estimating modality confidence but lack awareness of cross-modality alignment. Consequently, their performance is relatively stable under settings with only modality-specific noise (\(\eta = 0, \epsilon = 5\)), while their performance degrades significantly once cross-modality noise is introduced (all settings with \(\eta = 10\%\)). SMILE, SPS, ASL, and DisNCL primarily focus on cross-modality alignment while paying limited attention to confidence of each modality. As a result, they exhibit relatively stable performance when only cross-modality noise is present (\(\eta=10\%, \epsilon=0\)), but their performance degrades noticeably once modality-specific noise is introduced (all settings with \(\epsilon=5\)). In contrast, TAHCD consistently achieves more stable performance than the compared methods under various noise conditions on all datasets, demonstrating superior robustness. This advantage primarily stems from its explicit modeling of both modality-specific and cross-modality noise and their joint removal across multiple levels, enabling more thorough noise suppression.

\begin{table*}[t]
\caption{Comparison of classification performance between the proposed TAHCD and other reliable multimodal classification methods on the BRCA, ROSMAP, CUB, and UPMC FOOD101 datasets, where both training and testing are conducted under the same noise settings. The best results under various noise settings are indicated in bold, while the second-best results are underlined.}
\renewcommand{\arraystretch}{0.6}
\resizebox{\linewidth}{!}{%
\begin{tabular}{c|c|ccc|ccc|ccc|ccc}
\toprule
\multirow{2}{*}{Data Type}                                          & \multirow{2}{*}{Methods} & \multicolumn{3}{c|}{BRCA}                                            & \multicolumn{3}{c|}{ROSMAP}                              & \multicolumn{3}{c|}{CUB}                                             & \multicolumn{3}{c}{FOOD101}                                         \\
                                                                    &                          & \multicolumn{1}{c}{ACC} & \multicolumn{1}{c}{WeightedF1} & MacroF1 & \multicolumn{1}{c}{ACC} & \multicolumn{1}{c}{F1} & AUC & \multicolumn{1}{c}{ACC} & \multicolumn{1}{c}{WeightedF1} & MacroF1 & \multicolumn{1}{c}{ACC} & \multicolumn{1}{c}{WeightedF1} & MacroF1 \\ \midrule
\multirow{9}{*}{\begin{tabular}[c]{@{}c@{}}$\eta=0$\\ $\epsilon=0$\end{tabular}} & MD \cite{han2022multimodal}                      & \multicolumn{1}{c}{82.1}   & \multicolumn{1}{c}{81.7}          & 76.4       & \multicolumn{1}{c}{80.2}   & \multicolumn{1}{c}{82.9}  & 87.1   & \multicolumn{1}{c}{89.2}   & \multicolumn{1}{c}{89.1}          & 88.9       & \multicolumn{1}{c}{92.2}   & \multicolumn{1}{c}{92.4}          & 92.3       \\
                                                                    & MLCLNet \cite{zheng2023multi}                 & \multicolumn{1}{c}{82.3}   & \multicolumn{1}{c}{82.2}          & 78.6       & \multicolumn{1}{c}{79.4}   & \multicolumn{1}{c}{79.2}  & 89.3   & \multicolumn{1}{c}{90.2}   & \multicolumn{1}{c}{90.3}          & 89.9       & \multicolumn{1}{c}{92.1}   & \multicolumn{1}{c}{92.2}          & 92.1       \\ 
                                                                    & QMF \cite{zhang2023provable}                     & \multicolumn{1}{c}{82.5}   & \multicolumn{1}{c}{82.3}          & 79.1       & \multicolumn{1}{c}{78.3}   & \multicolumn{1}{c}{78.1}  & 85.1   & \multicolumn{1}{c}{88.3}   & \multicolumn{1}{c}{88.3}          & 87.9       & \multicolumn{1}{c}{91.7}   & \multicolumn{1}{c}{92.2}          & 92.1       \\ 
                                                                    & PDF \cite{cao2024predictive}                     & \multicolumn{1}{c}{\underline{83.7}}   & \multicolumn{1}{c}{\underline{83.6}}          & \underline{79.9}       & \multicolumn{1}{c}{\underline{83.0}}   & \multicolumn{1}{c}{\underline{83.0}}  & \underline{90.2}   & \multicolumn{1}{c}{\underline{91.9}}   & \multicolumn{1}{c}{\underline{92.0}}          & \underline{91.9}       & \multicolumn{1}{c}{\underline{92.3}}   & \multicolumn{1}{c}{\underline{92.7}}          & \underline{92.7}   \\ 
                                                                    \cmidrule{2-14}
                                                                
                                                                    & SMILE \cite{zeng2023semantic}                   & \multicolumn{1}{c}{80.5}   & \multicolumn{1}{c}{81.0}          & 78.2       & \multicolumn{1}{c}{79.2}   & \multicolumn{1}{c}{79.5}  & 87.4   & \multicolumn{1}{c}{88.6}   & \multicolumn{1}{c}{88.6}          & 88.3       & \multicolumn{1}{c}{91.3}   & \multicolumn{1}{c}{91.7}          & 91.6       \\ 
                                                                    & SPS \cite{xie2025seeking}                   & \multicolumn{1}{c}{79.8}   & \multicolumn{1}{c}{80.3}          & 78.4       & \multicolumn{1}{c}{80.4}   & \multicolumn{1}{c}{80.8}  & 88.1   & \multicolumn{1}{c}{89.3}   & \multicolumn{1}{c}{89.4}          & 89.1       & \multicolumn{1}{c}{91.9}   & \multicolumn{1}{c}{92.0}          & 91.8\\
                                                                    & ASL \cite{wang2025noisy}                     & \multicolumn{1}{c}{80.9}   & \multicolumn{1}{c}{80.5}          & 79.1       & \multicolumn{1}{c}{81.2}   & \multicolumn{1}{c}{81.0}  & 88.4   & \multicolumn{1}{c}{89.9}   & \multicolumn{1}{c}{89.7}          & 89.0       & \multicolumn{1}{c}{92.1}   & \multicolumn{1}{c}{92.5}          & 91.9       \\ 
                                                                    & DisNCL \cite{dang2025disentangled}                   & \multicolumn{1}{c}{81.2}   & \multicolumn{1}{c}{81.2}          & 79.3       & \multicolumn{1}{c}{81.0}   & \multicolumn{1}{c}{81.3}  & 88.5   & \multicolumn{1}{c}{90.0}   & \multicolumn{1}{c}{90.2}          & 90.2       & \multicolumn{1}{c}{92.2}   & \multicolumn{1}{c}{92.5}          & 92.4       \\ 
                                                                    \cmidrule{2-14} 
                                                                    & TAHCD                  & \multicolumn{1}{c}{\textbf{85.6}}   & \multicolumn{1}{c}{\textbf{85.7}}          & \textbf{82.4}       & \multicolumn{1}{c}{\textbf{87.7}}   & \multicolumn{1}{c}{\textbf{88.1}}  & \textbf{93.2}   & \multicolumn{1}{c}{\textbf{93.8}}   & \multicolumn{1}{c}{\textbf{93.8}}          & \textbf{93.6}       & \multicolumn{1}{c}{\textbf{94.2}}   & \multicolumn{1}{c}{\textbf{94.4}}          & \textbf{94.2}       \\ \midrule 
                                                                    
\multirow{9}{*}{\begin{tabular}[c]{@{}c@{}}$\eta=0$\\ $\epsilon=5$\end{tabular}} & MD \cite{han2022multimodal}                      & \multicolumn{1}{c}{76.0}   & \multicolumn{1}{c}{73.6}          & 67.4       & \multicolumn{1}{c}{68.9}   & \multicolumn{1}{c}{73.6}  & 78.6   & \multicolumn{1}{c}{88.2}   & \multicolumn{1}{c}{88.2}          & 88.2       & \multicolumn{1}{c}{89.2}   & \multicolumn{1}{c}{90.1}          & 90.1      \\  
                                                                    & MLCLNet \cite{zheng2023multi}                & \multicolumn{1}{c}{77.0}   & \multicolumn{1}{c}{76.6}          & 70.8       & \multicolumn{1}{c}{74.1}   & \multicolumn{1}{c}{78.5}  & 82.5   & \multicolumn{1}{c}{86.4}   & \multicolumn{1}{c}{86.5}          & 86.3       & \multicolumn{1}{c}{88.7}   & \multicolumn{1}{c}{88.5}          & 88.7      \\ 
                                                                    & QMF \cite{zhang2023provable}                     & \multicolumn{1}{c}{76.4}   & \multicolumn{1}{c}{75.8}          & \underline{71.0}       & \multicolumn{1}{c}{70.8}   & \multicolumn{1}{c}{75.0}  & 78.9   & \multicolumn{1}{c}{86.5}   & \multicolumn{1}{c}{86.5}          & 86.4       & \multicolumn{1}{c}{91.2}   & \multicolumn{1}{c}{91.7}          & 91.7       \\ 
                                                                    & PDF \cite{cao2024predictive}                    & \multicolumn{1}{c}{\underline{78.3}}   & \multicolumn{1}{c}{\underline{77.4}}          & 69.0       & \multicolumn{1}{c}{\underline{75.5}}   & \multicolumn{1}{c}{\underline{79.0}}  & \underline{83.2}   & \multicolumn{1}{c}{\underline{90.1}}   & \multicolumn{1}{c}{\underline{90.1}}          & \underline{90.2}       & \multicolumn{1}{c}{\underline{92.2}}   & \multicolumn{1}{c}{\underline{92.5}}          & \underline{92.4}      \\
                                                                    \cmidrule{2-14}
                                                                    & SMILE \cite{zeng2023semantic}                   & \multicolumn{1}{c}{71.5}   & \multicolumn{1}{c}{67.0}          & 54.3       & \multicolumn{1}{c}{67.6}   & \multicolumn{1}{c}{72.3}  & 71.1   & \multicolumn{1}{c}{82.0}   & \multicolumn{1}{c}{81.9}          & 81.9       & \multicolumn{1}{c}{89.7}   & \multicolumn{1}{c}{90.1}          & 90.0       \\ 
                                                                    & SPS \cite{xie2025seeking}                   & \multicolumn{1}{c}{72.1}   & \multicolumn{1}{c}{67.4}          & 54.7       & \multicolumn{1}{c}{70.2}   & \multicolumn{1}{c}{74.5}  & 75.2   & \multicolumn{1}{c}{83.4}   & \multicolumn{1}{c}{83.2}          & 83.6       & \multicolumn{1}{c}{90.5}   & \multicolumn{1}{c}{90.7}          & 90.5       \\
                                                                    & ASL \cite{wang2025noisy}                     & \multicolumn{1}{c}{75.5}   & \multicolumn{1}{c}{73.2}          & 69.1       & \multicolumn{1}{c}{73.4}   & \multicolumn{1}{c}{74.1}  & 81.5   & \multicolumn{1}{c}{85.1}   & \multicolumn{1}{c}{85.0}          & 85.2       & \multicolumn{1}{c}{91.1}   & \multicolumn{1}{c}{91.5}          & 91.5       \\ 
                                                                    & DisNCL \cite{dang2025disentangled}                   & \multicolumn{1}{c}{76.3}   & \multicolumn{1}{c}{73.4}          & 69.8       & \multicolumn{1}{c}{74.2}   & \multicolumn{1}{c}{74.6}  & 81.9   & \multicolumn{1}{c}{85.2}   & \multicolumn{1}{c}{85.3}          & 85.3       & \multicolumn{1}{c}{91.2}   & \multicolumn{1}{c}{91.5}          & 91.4       \\ 
                                                                    \cmidrule{2-14} 
                                                                    & TAHCD                  & \multicolumn{1}{c}{\textbf{84.4}}   & \multicolumn{1}{c}{\textbf{84.0}}          & \textbf{80.0}       & \multicolumn{1}{c}{\textbf{81.1}}   & \multicolumn{1}{c}{\textbf{81.1}}  & \textbf{85.6}   & \multicolumn{1}{c}{\textbf{93.6}}   & \multicolumn{1}{c}{\textbf{93.5}}          & \textbf{93.5}       & \multicolumn{1}{c}{\textbf{94.0}}   & \multicolumn{1}{c}{\textbf{94.1}}          & \textbf{94.1}       \\ \midrule 
                                                                    
\multirow{9}{*}{\begin{tabular}[c]{@{}c@{}}$\eta=10\%$\\ $\epsilon=0$\end{tabular}} & MD \cite{han2022multimodal}                      & \multicolumn{1}{c}{65.0}   & \multicolumn{1}{c}{56.7}          & 41.7       & \multicolumn{1}{c}{67.9}   & \multicolumn{1}{c}{72.8}  & 77.3   & \multicolumn{1}{c}{69.4}   & \multicolumn{1}{c}{69.3}          & 69.0       & \multicolumn{1}{c}{73.1}   & \multicolumn{1}{c}{73.2}          & 73.2       \\ 
                                                                    & MLCLNet \cite{zheng2023multi}                 & \multicolumn{1}{c}{70.4}   & \multicolumn{1}{c}{68.2}          & 50.5       & \multicolumn{1}{c}{70.4}   & \multicolumn{1}{c}{70.6}  & 76.2   & \multicolumn{1}{c}{69.2}   & \multicolumn{1}{c}{68.8}          & 68.1       & \multicolumn{1}{c}{73.2}   & \multicolumn{1}{c}{73.0}          & 73.1      \\ 
                                                                    & QMF \cite{zhang2023provable}                      & \multicolumn{1}{c}{68.8}   & \multicolumn{1}{c}{64.9}          & 50.3       & \multicolumn{1}{c}{70.8}   & \multicolumn{1}{c}{69.3}  & 73.1   & \multicolumn{1}{c}{70.3}   & \multicolumn{1}{c}{70.2}          & 69.5       & \multicolumn{1}{c}{74.4}   & \multicolumn{1}{c}{74.3}         & 74.3       \\ 
                                                                    & PDF \cite{cao2024predictive}                      & \multicolumn{1}{c}{70.8}   & \multicolumn{1}{c}{69.7}          & 54.5       & \multicolumn{1}{c}{72.6}   & \multicolumn{1}{c}{73.9}  & 78.1   & \multicolumn{1}{c}{69.4}   & \multicolumn{1}{c}{69.3}          & 68.3       & \multicolumn{1}{c}{73.1}   & \multicolumn{1}{c}{73.2}          & 73.2   \\ 
                                                                    \cmidrule{2-14} 
                                                                    & SMILE \cite{zeng2023semantic}                    & \multicolumn{1}{c}{68.9}   & \multicolumn{1}{c}{64.6}          & 50.2       & \multicolumn{1}{c}{69.1}   & \multicolumn{1}{c}{70.0}  & 77.4   & \multicolumn{1}{c}{62.2}   & \multicolumn{1}{c}{62.3}          & 61.9       & \multicolumn{1}{c}{73.2}   & \multicolumn{1}{c}{73.4}          & 73.4       \\ 
                                                                    & SPS \cite{xie2025seeking}                    & \multicolumn{1}{c}{71.6}   & \multicolumn{1}{c}{71.1}          & 58.2       & \multicolumn{1}{c}{74.2}   & \multicolumn{1}{c}{75.5}  & 81.3   & \multicolumn{1}{c}{71.6}   & \multicolumn{1}{c}{71.2}          & 71.9       & \multicolumn{1}{c}{74.1}   & \multicolumn{1}{c}{74.4}          & 74.4       \\
                                                                    & ASL \cite{wang2025noisy}                     & \multicolumn{1}{c}{71.9}   & \multicolumn{1}{c}{71.4}          & 62.8       & \multicolumn{1}{c}{74.7}   & \multicolumn{1}{c}{76.2}  & 82.3   & \multicolumn{1}{c}{71.7}   & \multicolumn{1}{c}{71.6}          & 71.6       & \multicolumn{1}{c}{74.2}   & \multicolumn{1}{c}{74.3}          & 74.3       \\ 
                                                                    & DisNCL \cite{dang2025disentangled}                   & \multicolumn{1}{c}{\underline{72.0}}   & \multicolumn{1}{c}{\underline{72.3}}          & \underline{62.6}       & \multicolumn{1}{c}{\underline{75.1}}   & \multicolumn{1}{c}{\underline{76.3}}  & \underline{82.4}   & \multicolumn{1}{c}{\underline{71.9}}   & \multicolumn{1}{c}{\underline{71.8}}          & \underline{72.0}       & \multicolumn{1}{c}{\underline{74.8}}   & \multicolumn{1}{c}{\underline{74.6}}          & \underline{74.5}       \\ 
                                                                    \cmidrule{2-14} 
                                                                    & TAHCD                  & \multicolumn{1}{c}{\textbf{77.4}}   & \multicolumn{1}{c}{\textbf{77.0}}          & \textbf{69.4}       & \multicolumn{1}{c}{\textbf{79.2}}   & \multicolumn{1}{c}{\textbf{79.6}}  & \textbf{84.5}   & \multicolumn{1}{c}{\textbf{76.4}}   & \multicolumn{1}{c}{\textbf{76.3}}          & \textbf{76.6}       & \multicolumn{1}{c}{\textbf{77.5}}   & \multicolumn{1}{c}{\textbf{77.6}}          & \textbf{77.6}      \\ \midrule 
                                                                    
\multirow{9}{*}{\begin{tabular}[c]{@{}c@{}}$\eta=10\%$\\ $\epsilon=5$ \end{tabular}} & MD \cite{han2022multimodal}                      & \multicolumn{1}{c}{60.5}   & \multicolumn{1}{c}{48.1}          & 31.0       & \multicolumn{1}{c}{66.0}   & \multicolumn{1}{c}{72.7}  & 72.1   & \multicolumn{1}{c}{68.5}   & \multicolumn{1}{c}{68.0}          & 67.8      & \multicolumn{1}{c}{69.2}   & \multicolumn{1}{c}{69.3}          & 69.3      \\  
                                                                    & MLCLNet \cite{zheng2023multi}                & \multicolumn{1}{c}{65.1}   & \multicolumn{1}{c}{58.2}          & 42.1       & \multicolumn{1}{c}{64.9}   & \multicolumn{1}{c}{70.1}  & 71.5   & \multicolumn{1}{c}{67.6}   & \multicolumn{1}{c}{67.7}          & 68.3       & \multicolumn{1}{c}{69.9}   & \multicolumn{1}{c}{69.8}          & 70.1    \\ 
                                                                    & QMF \cite{zhang2023provable}                      & \multicolumn{1}{c}{65.0}   & \multicolumn{1}{c}{58.9}          & 44.4       & \multicolumn{1}{c}{62.2}   & \multicolumn{1}{c}{69.2}  & 70.1   & \multicolumn{1}{c}{64.8}   & \multicolumn{1}{c}{65.1}          & 64.8       & \multicolumn{1}{c}{71.2}   & \multicolumn{1}{c}{71.7}          & 71.7       \\ 
                                                                    & PDF \cite{cao2024predictive}                      & \multicolumn{1}{c}{65.8}   & \multicolumn{1}{c}{58.5}          & 42.4       & \multicolumn{1}{c}{68.9}   & \multicolumn{1}{c}{72.7}  & 72.3   & \multicolumn{1}{c}{68.5}   & \multicolumn{1}{c}{67.9}          & 67.2       & \multicolumn{1}{c}{\underline{72.3}}   & \multicolumn{1}{c}{\underline{72.4}}          & \underline{72.4}   \\
                                                                    \cmidrule{2-14} 
                                                                    & SMILE \cite{zeng2023semantic}                   & \multicolumn{1}{c}{66.8}   & \multicolumn{1}{c}{65.9}          & 47.8       & \multicolumn{1}{c}{68.4}   & \multicolumn{1}{c}{68.9}  & 70.2   & \multicolumn{1}{c}{56.8}   & \multicolumn{1}{c}{56.4}          & 55.9       & \multicolumn{1}{c}{70.8}   & \multicolumn{1}{c}{71.3}          & 71.2       \\ 
                                                                    & SPS \cite{xie2025seeking}                   & \multicolumn{1}{c}{68.1}   & \multicolumn{1}{c}{66.2}          & 48.3       & \multicolumn{1}{c}{69.2}   & \multicolumn{1}{c}{72.4}  & 71.3   & \multicolumn{1}{c}{70.2}   & \multicolumn{1}{c}{70.1}          & \underline{70.9}       & \multicolumn{1}{c}{71.8}   & \multicolumn{1}{c}{71.8}          & 71.5       \\
                                                                    & ASL \cite{wang2025noisy}                     & \multicolumn{1}{c}{68.2}   & \multicolumn{1}{c}{66.5}          & 50.2       & \multicolumn{1}{c}{70.4}   & \multicolumn{1}{c}{73.1}  & 71.8   & \multicolumn{1}{c}{70.4}   & \multicolumn{1}{c}{70.5}          & 70.5       & \multicolumn{1}{c}{72.0}   & \multicolumn{1}{c}{71.8}          & 71.7       \\ 
                                                                    & DisNCL \cite{dang2025disentangled}                   & \multicolumn{1}{c}{\underline{68.5}}   & \multicolumn{1}{c}{\underline{67.1}}          & \underline{50.9}       & \multicolumn{1}{c}{\underline{70.9}}   & \multicolumn{1}{c}{\underline{73.3}}  & \underline{72.5}   & \multicolumn{1}{c}{\underline{70.8}}   & \multicolumn{1}{c}{\underline{70.9}}          & \underline{70.9}       & \multicolumn{1}{c}{72.2}   & \multicolumn{1}{c}{72.3}          & 72.3       \\ 
                                                                    \cmidrule{2-14} 
                                                                    & TAHCD                  & \multicolumn{1}{c}{\textbf{73.2}}   & \multicolumn{1}{c}{\textbf{71.9}}          & \textbf{60.1}       & \multicolumn{1}{c}{\textbf{76.4}}   & \multicolumn{1}{c}{\textbf{78.6}}  & \textbf{77.2}   & \multicolumn{1}{c}{\textbf{75.3}}   & \multicolumn{1}{c}{\textbf{75.2}}          & \textbf{75.4}       & \multicolumn{1}{c}{\textbf{76.3}}   & \multicolumn{1}{c}{\textbf{76.1}}          & \textbf{76.2}       \\ \bottomrule
\end{tabular}%
}
\label{tab:sota}
\end{table*}
\subsubsection{Generalization ability} To evaluate the improved generalization ability of TAHCD under previously unseen noise, we train all methods on the original datasets and test them on noise-corrupted data. Table \ref{tab:sota-gen} reports the results on test data corrupted by a mixture of modality-specific noise with \(\epsilon=5\) and cross-modality noise with \(\eta=10\%\). When confronted with noise unseen during training, all baseline methods suffer substantial performance degradation. Specifically, compared with the results under the noise setting of \(\eta=10\%\) and \(\epsilon=5\) in Table \ref{tab:sota} (where noise is added on both train and test sets), their performance in Table \ref{tab:sota-gen} degrades substantially, indicating limited adaptability to previously unseen noise during training. In contrast, our TAHCD outperforms other baselines when exposed to previously unseen noise, while exhibiting a noticeably smaller performance degradation compared to the setting where such noise is observed during training. The possible reason is that TAHCD can better adapt in response to input noise at test time, thereby exhibiting stronger generalization ability.

\begin{table*}[t]
\caption{Comparison of classification performance between the proposed TAHCD and other reliable multimodal classification methods on the BRCA, ROSMAP, CUB, and UPMC FOOD101 datasets. All methods are trained on the original datasets and test on data corrupted with multimodal noise  $(\eta=10\%,\epsilon=5)$.}
\renewcommand{\arraystretch}{0.6}
\resizebox{\linewidth}{!}{%
\begin{tabular}{c|ccc|ccc|ccc|ccc}
\toprule
\multirow{2}{*}{Methods} & \multicolumn{3}{c|}{BRCA}                                            & \multicolumn{3}{c|}{ROSMAP}                              & \multicolumn{3}{c|}{CUB}                                             & \multicolumn{3}{c}{FOOD101}                                         \\  
                                                                    & \multicolumn{1}{c}{ACC} & \multicolumn{1}{c}{WeightedF1} & MacroF1 & \multicolumn{1}{c}{ACC} & \multicolumn{1}{c}{F1} & AUC & \multicolumn{1}{c}{ACC} & \multicolumn{1}{c}{WeightedF1} & MacroF1 & \multicolumn{1}{c}{ACC} & \multicolumn{1}{c}{WeightedF1} & MacroF1 \\ \midrule
 MD \cite{han2022multimodal}                      & \multicolumn{1}{c}{51.4}   & \multicolumn{1}{c}{50.8}          & 44.3       & \multicolumn{1}{c}{58.6}   & \multicolumn{1}{c}{58.5}  & 57.3   & \multicolumn{1}{c}{57.4}   & \multicolumn{1}{c}{57.7}          & 57.6       & \multicolumn{1}{c}{58.1}   & \multicolumn{1}{c}{57.8}          & 57.8   \\ 
 MLCLNet \cite{zheng2023multi}                 & \multicolumn{1}{c}{49.6}   & \multicolumn{1}{c}{48.7}          & 42.1       & \multicolumn{1}{c}{56.7}   & \multicolumn{1}{c}{56.1}  & 54.8   & \multicolumn{1}{c}{54.8}   & \multicolumn{1}{c}{54.9}          & 54.9       & \multicolumn{1}{c}{55.4}   & \multicolumn{1}{c}{55.7}          & 55.7      \\ 
 QMF \cite{zhang2023provable}                     & \multicolumn{1}{c}{56.2}   & \multicolumn{1}{c}{55.8}          & 49.3       & \multicolumn{1}{c}{64.4}   & \multicolumn{1}{c}{64.6}  & 65.5   & \multicolumn{1}{c}{61.5}   & \multicolumn{1}{c}{61.3}          & 61.1       & \multicolumn{1}{c}{63.6}   & \multicolumn{1}{c}{63.7}          & 63.6       \\ 
PDF \cite{cao2024predictive}                     & \multicolumn{1}{c}{56.6}   & \multicolumn{1}{c}{56.4}          & 49.9       & \multicolumn{1}{c}{65.1}   & \multicolumn{1}{c}{65.3}  & 66.3   & \multicolumn{1}{c}{62.2}   & \multicolumn{1}{c}{62.3}          & 62.4       & \multicolumn{1}{c}{64.4}   & \multicolumn{1}{c}{64.8}          & 64.7       \\ 
SMILE \cite{zeng2023semantic}                   & \multicolumn{1}{c}{45.0}   & \multicolumn{1}{c}{44.1}          & 38.7       & \multicolumn{1}{c}{52.8}   & \multicolumn{1}{c}{52.8}  & 53.5   & \multicolumn{1}{c}{49.2}   & \multicolumn{1}{c}{49.1}          & 49.1       & \multicolumn{1}{c}{50.7}   & \multicolumn{1}{c}{50.8}          & 50.8       \\
SPS \cite{xie2025seeking}                   & \multicolumn{1}{c}{46.1}   & \multicolumn{1}{c}{45.3}          & 40.1       & \multicolumn{1}{c}{56.6}   & \multicolumn{1}{c}{56.9}  & 57.1   & \multicolumn{1}{c}{52.1}   & \multicolumn{1}{c}{52.1}          & 52.7       & \multicolumn{1}{c}{55.9}   & \multicolumn{1}{c}{56.1}          & 56.0       \\
ASL \cite{wang2025noisy}                     & \multicolumn{1}{c}{45.8}   & \multicolumn{1}{c}{44.6}          & 39.2       & \multicolumn{1}{c}{56.2}   & \multicolumn{1}{c}{56.0}  & 57.3   & \multicolumn{1}{c}{51.5}   & \multicolumn{1}{c}{51.4}          & 51.2       & \multicolumn{1}{c}{54.4}   & \multicolumn{1}{c}{54.3}          & 54.3       \\ 
DisNCL \cite{dang2025disentangled}                   & \multicolumn{1}{c}{46.5}   & \multicolumn{1}{c}{45.4}          & 40.0       & \multicolumn{1}{c}{56.3}   & \multicolumn{1}{c}{56.2}  & 57.6   & \multicolumn{1}{c}{52.2}   & \multicolumn{1}{c}{52.2}          & 52.1       & \multicolumn{1}{c}{55.3}   & \multicolumn{1}{c}{55.1}          & 55.2       \\
\midrule
TAHCD                  & \multicolumn{1}{c}{\textbf{67.4}}   & \multicolumn{1}{c}{\textbf{66.2}}          & \textbf{59.2}       & \multicolumn{1}{c}{\textbf{75.1}}   & \multicolumn{1}{c}{\textbf{77.3}}  & \textbf{76.4}   & \multicolumn{1}{c}{\textbf{72.4}}   & \multicolumn{1}{c}{\textbf{72.4}}          & \textbf{71.8}       & \multicolumn{1}{c}{\textbf{74.2}}   & \multicolumn{1}{c}{\textbf{74.1}}          & \textbf{74.1}       \\  \bottomrule
\end{tabular}%
}
\label{tab:sota-gen}
\end{table*}

\begin{table}[t]
\centering
\caption{Ablation studies on four datasets under noise ($\eta=10\%,\epsilon=5$).}

\begin{minipage}{0.98\linewidth}
    \scriptsize  
    \setlength{\tabcolsep}{3pt}  
    
    \begin{subtable}[t]{0.47\linewidth} 
      \raggedright 
      \renewcommand{\arraystretch}{0.6}
      \caption{Noise added on both training and testing data.}
      \resizebox{\linewidth}{!}{
      \begin{tabular}{cccc|ccc}
\toprule
Dataset & ASSA & SACA & TTCE & ACC & WeightedF1 & MacroF1 \\ \midrule
\multirow{4}{*}{BRCA} & \ding{55}                     & \ding{55}                     & \ding{55}                     & \multicolumn{1}{c}{48.3}   & \multicolumn{1}{c}{47.1}       & 45.5          \\ 
& \ding{51}                     & \ding{55}                     & \ding{55}                    & \multicolumn{1}{c}{65.8}   & \multicolumn{1}{c}{64.0}       & 56.5          \\
& \ding{51}                     & \ding{51}                     & \ding{55}                     & \multicolumn{1}{c}{67.4}   & \multicolumn{1}{c}{65.6}       & 58.4          \\
& \ding{51}                     & \ding{51}                     & \ding{51}                     & \multicolumn{1}{c}{\textbf{73.2}}   & \multicolumn{1}{c}{\textbf{71.9}}       & \textbf{60.1}          \\ 
\midrule
\multirow{4}{*}{CUB} & \ding{55}                     & \ding{55}                     & \ding{55}                     & \multicolumn{1}{c}{58.6}   & \multicolumn{1}{c}{58.4}       & 58.6          \\ 
& \ding{51}                     & \ding{55}                     & \ding{55}                    & \multicolumn{1}{c}{69.8}   & \multicolumn{1}{c}{69.7}       & 69.7          \\
& \ding{51}                     & \ding{51}                     & \ding{55}                     & \multicolumn{1}{c}{71.3}   & \multicolumn{1}{c}{71.3}       & 71.0          \\ 
& \ding{51}                     & \ding{51}                     & \ding{51}                     & \multicolumn{1}{c}{\textbf{75.3}}   & \multicolumn{1}{c}{\textbf{75.2}}       & \textbf{75.4}          \\ 
\midrule
\multirow{4}{*}{FOOD101} & \ding{55}                     & \ding{55}                     & \ding{55}                     & \multicolumn{1}{c}{60.8}   & \multicolumn{1}{c}{60.6}       & 60.9          \\ 
& \ding{51}                     & \ding{55}                     & \ding{55}                    & \multicolumn{1}{c}{72.3}   & \multicolumn{1}{c}{72.4}       & 72.3          \\
& \ding{51}                     & \ding{51}                     & \ding{55}                     & \multicolumn{1}{c}{74.7}   & \multicolumn{1}{c}{74.6}       & 74.7          \\
& \ding{51}                     & \ding{51}                     & \ding{51}                     & \multicolumn{1}{c}{\textbf{76.3}}   & \multicolumn{1}{c}{\textbf{76.1}}       & \textbf{76.2}          \\ 

\midrule

Dataset & ASSA & SACA & TTCE & ACC & F1 & AUC \\ \midrule
\multirow{4}{*}{ROSMAP} & \ding{55}                     & \ding{55}                     & \ding{55}                     & \multicolumn{1}{c}{53.1}   & \multicolumn{1}{c}{52.7}       & 53.8          \\
& \ding{51}                     & \ding{55}                     & \ding{55}                    & \multicolumn{1}{c}{72.7}   & \multicolumn{1}{c}{73.8}       & 74.0          \\ 
& \ding{51}                     & \ding{51}                     & \ding{55}                     & \multicolumn{1}{c}{74.6}   & \multicolumn{1}{c}{76.1}       & 76.2          \\
& \ding{51}                     & \ding{51}                     & \ding{51}                     & \multicolumn{1}{c}{\textbf{76.4}}   & \multicolumn{1}{c}{\textbf{78.6}}       & \textbf{77.2}          \\ 
 
\bottomrule
\end{tabular}
}
\label{tab:abl-rob}
    \end{subtable}
    \hspace{0.04\linewidth}  
    \begin{subtable}[t]{0.47\linewidth}
      \raggedright 
      \caption{Noise only added on testing data.}
      \renewcommand{\arraystretch}{0.6}
      \resizebox{\linewidth}{!}{
      \begin{tabular}{cccc|ccc}
\toprule
Dataset & ASSA & SACA & TTCE & ACC & WeightedF1 & MacroF1 \\ \midrule
\multirow{4}{*}{BRCA} & \ding{55}                     & \ding{55}                     & \ding{55}                     & \multicolumn{1}{c}{46.5}   & \multicolumn{1}{c}{46.7}       & 43.1          \\ 
& \ding{51}                     & \ding{55}                     & \ding{55}                    & \multicolumn{1}{c}{49.2}   & \multicolumn{1}{c}{49.0}       & 45.4         \\
& \ding{51}                     & \ding{51}                     & \ding{55}                     & \multicolumn{1}{c}{50.1}   & \multicolumn{1}{c}{49.8}       & 47.2          \\
& \ding{51}                     & \ding{51}                     & \ding{51}                     & \multicolumn{1}{c}{\textbf{67.4}}   & \multicolumn{1}{c}{\textbf{66.2}}       & \textbf{59.2}          \\ 
\midrule
\multirow{4}{*}{CUB} & \ding{55}                     & \ding{55}                     & \ding{55}                     & \multicolumn{1}{c}{51.6}   & \multicolumn{1}{c}{51.4}       & 51.6          \\
& \ding{51}                     & \ding{55}                     & \ding{55}                    & \multicolumn{1}{c}{53.3}   & \multicolumn{1}{c}{53.3}       & 52.7          \\
& \ding{51}                     & \ding{51}                     & \ding{55}                     & \multicolumn{1}{c}{54.6}   & \multicolumn{1}{c}{54.6}       & 53.8          \\ 
& \ding{51}                     & \ding{51}                     & \ding{51}                     & \multicolumn{1}{c}{\textbf{72.4}}   & \multicolumn{1}{c}{\textbf{72.4}}       & \textbf{71.8}          \\ 
\midrule
\multirow{4}{*}{FOOD101} & \ding{55}                     & \ding{55}                     & \ding{55}                     & \multicolumn{1}{c}{51.9}   & \multicolumn{1}{c}{51.8}       & 52.1          \\
& \ding{51}                     & \ding{55}                     & \ding{55}                    & \multicolumn{1}{c}{54.3}   & \multicolumn{1}{c}{54.3}       & 54.3          \\
& \ding{51}                     & \ding{51}                     & \ding{55}                     & \multicolumn{1}{c}{55.6}   & \multicolumn{1}{c}{54.9}       & 55.1          \\
& \ding{51}                     & \ding{51}                     & \ding{51}                     & \multicolumn{1}{c}{\textbf{74.2}}   & \multicolumn{1}{c}{\textbf{74.1}}       & \textbf{74.1}          \\ 

\midrule

Dataset & ASSA & SACA & TTCE & ACC & F1 & AUC \\ \midrule
\multirow{4}{*}{ROSMAP} & \ding{55}                     & \ding{55}                     & \ding{55}                     & \multicolumn{1}{c}{52.2}   & \multicolumn{1}{c}{52.3}       & 50.8          \\
& \ding{51}                     & \ding{55}                     & \ding{55}                    & \multicolumn{1}{c}{54.8}   & \multicolumn{1}{c}{55.9}       & 54.8         \\ 
& \ding{51}                     & \ding{51}                     & \ding{55}                     & \multicolumn{1}{c}{56.3}   & \multicolumn{1}{c}{57.4}       & 55.7          \\ 
& \ding{51}                     & \ding{51}                     & \ding{51}                     & \multicolumn{1}{c}{\textbf{75.1}}   & \multicolumn{1}{c}{\textbf{77.3}}       & \textbf{76.4}          \\ 

\bottomrule
\end{tabular}
}
\label{tab:abl-gen}
    \end{subtable}
    \setlength{\tabcolsep}{6pt}
  \end{minipage}

\end{table}

\subsection{Abliation Study}\label{sec:abl-study}
To evaluate the overall effectiveness of the three key components of TAHCD, namely Adaptive Stable Subspace Alignment (ASSA), Sample-Adaptive Confidence Alignment (SACA), and Test-Time Cooperative Enhancement (TTCE), we conduct ablation studies. As shown in Table \ref{tab:abl-rob}, following the same setting as Table \ref{tab:sota}, identical noise is added to both the training and test data to assess the contribution of each component to model robustness. The results indicate that all three components contribute to robustness improvement, with ASSA providing the largest gain, followed by SACA and TTCE. A possible reason is that the noise is sampled from a certain distribution and added to all samples, exhibiting strong global characteristics, which allows ASSA to remove most of the noise at the global level. SACA also contributes to performance improvements, indicating that even noise sampled from the same distribution contains outlier components, thereby demonstrating the necessity of instance-level denoising on each sample. The performance gains brought by TTCE indicate that the cooperative enhancement between global-level and instance-level denoising is beneficial for improving robustness.

Additionally, to evaluate the contribution of each component to generalization ability, we follow the same setting as in Table \ref{tab:sota-gen} by training the model on the original dataset and testing it on noise-corrupted data. The results are shown in Table \ref{tab:abl-gen}. Removing TTCE leads to a substantial performance drop, indicating its critical role in adapting to previously unseen noise and improving generalization. When TTCE is removed, further removing ASSA or SACA results in additional performance degradation, indicating that these two components also contribute to learning more generalizable multimodal representations. A possible reason is that ASSA and SACA learn informative modality distribution priors, which help guide the model to better remove unseen noise.

\begin{figure}[t]
\centering
\includegraphics[width=\linewidth]{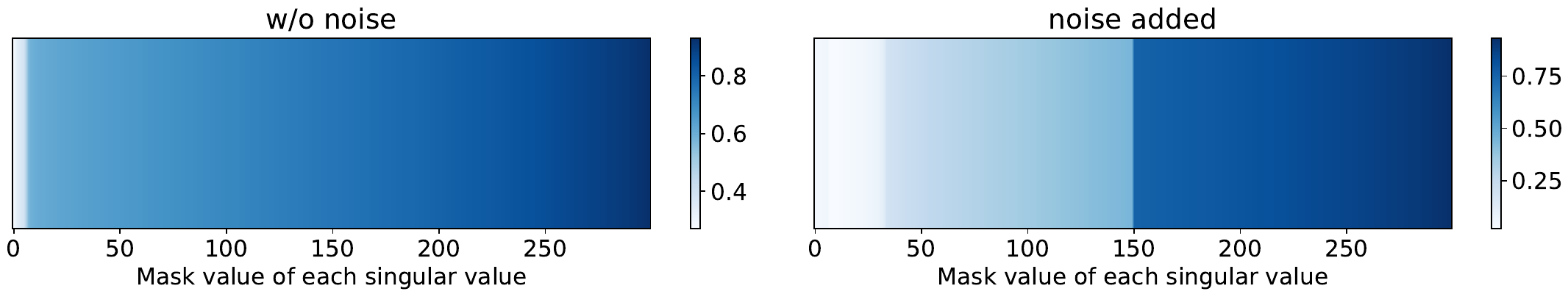} 
\caption{Visualization of masks \(w^m_{\lambda}\) learnd by ASSA on CUB image features, before and after adding noise (\(\epsilon=5\)) to half of the feature dimensions.}
\label{fig:svd-mask}
\end{figure}

\subsection{Discussion On ASSA}\label{sec:dis-assa}
As detailed in Sec. \ref{sec:assa}, ASSA learns a mask $w_{\lambda}^m$ based on the singular values obtained from the decomposition of modality feature covariance matrices to construct stable feature subspace, guided by two constraints \(\mathcal{L}_{\mathrm{o}}\) and \(\mathcal{L}_{\mathrm{a}}\).
\subsubsection{Discussion On Adaptive Stable Subspace Construction} 
We evaluate the effectiveness of stable subspace construction by examining whether ASSA learns noise-related masks \(w^m_{\lambda}\). Specifically, we generate Gaussian noise \(n^m \in \mathbb{R}^{N \times d^m}\) with severity \(\epsilon = 5\), set the values of half of the feature dimensions in the noise matrix to zero, and add the resulting noise to one modality \(x^m\). We then visualize the learned mask \(w^m_{\lambda}\) before and after noise corruption in Figure \ref{fig:svd-mask}, with results shown for the corrupted image modality on the CUB dataset. Before adding noise (left panel of Figure \ref{fig:svd-mask}), the learned mask show small values in few feature dimensions, suggesting that most features of the original high-quality data are informative and largely retained. After the noise is added (right panel of Figure \ref{fig:svd-mask}), approximately half of the learned mask values corresponding to corrupted feature dimensions decrease, appearing lighter in the figure. This indicates that ASSA can learn masks highly correlated with noise, thereby preserving informative principal axes and ensuring the stability of the constructed feature subspace.

\begin{table}[t]
\centering
\caption{Evaluation of the effects of \(\mathcal{L}_{\mathrm{o}}\) and \(\mathcal{L}_{\mathrm{a}}\) on datasets under noise.}

\begin{minipage}{0.98\linewidth}
    \scriptsize  
    \setlength{\tabcolsep}{3pt}  
    
    \begin{subtable}[t]{0.47\linewidth} 
      \raggedright  
      \caption{Under modality-specific noise (\(\epsilon=5\)).}
      \renewcommand{\arraystretch}{0.6}
\resizebox{\linewidth}{!}{
\begin{tabular}{ccccc}
\toprule
            Dataset   & Method  & ACC & WeightedF1 & MacroF1 \\ \midrule
\multirow{3}{*}{BRCA} & w/o $\mathcal{L}_{\mathrm{o}}$ & 75.1 & 74.8  & 70.3 \\ 
                  & w/o $\mathcal{L}_{\mathrm{a}}$ & 83.8 & 83.2 & 79.6 \\
                  & Proposed & \textbf{84.4} & \textbf{84.0} & \textbf{80.0} \\ \midrule
\multirow{3}{*}{CUB} & w/o $\mathcal{L}_{\mathrm{o}}$ & 84.4 & 84.3 & 84.4 \\ 
                  & w/o $\mathcal{L}_{\mathrm{a}}$ & 92.4 & 92.3 & 92.3 \\ 
                  & Proposed & \textbf{93.6} & \textbf{93.6} & \textbf{93.5} \\ \midrule
\multirow{3}{*}{FOOD101} & w/o $\mathcal{L}_{\mathrm{o}}$ & 87.1 & 87.0 & 87.0 \\ 
                  & w/o $\mathcal{L}_{\mathrm{a}}$ & 92.5 & 92.4 & 92.6 \\ 
                  & Proposed & \textbf{94.0} & \textbf{94.0} & \textbf{94.1} \\ \midrule
            Dataset   & Method  & ACC & F1 & AUC \\ \midrule
\multirow{3}{*}{ROSMAP} & w/o $\mathcal{L}_{\mathrm{o}}$ & 70.4 & 70.1 & 72.3 \\ 
                  & w/o $\mathcal{L}_{\mathrm{a}}$ & 79.8 & 79.6 & 84.1 \\
                  & Proposed & \textbf{81.1} & \textbf{81.1} & \textbf{85.6} \\ \bottomrule
\end{tabular}
}
\label{tab:contraint-ms}
    \end{subtable}
    \hspace{0.04\linewidth}  
    \begin{subtable}[t]{0.47\linewidth}
      \raggedright 
      \caption{Under cross-modality noise (\(\eta=10\%\)).}
      \renewcommand{\arraystretch}{0.6}
\resizebox{\linewidth}{!}{
\begin{tabular}{ccccc}
\toprule
            Dataset   & Method  & ACC & WeightedF1 & MacroF1 \\ \midrule
\multirow{3}{*}{BRCA} & w/o $\mathcal{L}_{\mathrm{o}}$ & 71.5 & 71.2  & 70.3 \\ 
                  & w/o $\mathcal{L}_{\mathrm{a}}$ & 59.7 & 59.1 & 57.8 \\
                  & Proposed & \textbf{77.4} & \textbf{77.0} & \textbf{69.4} \\ \midrule
\multirow{3}{*}{CUB} & w/o $\mathcal{L}_{\mathrm{o}}$ & 71.3 & 71.6 & 71.6 \\ 
                  & w/o $\mathcal{L}_{\mathrm{a}}$ & 60.1 & 60.3& 60.2 \\ 
                  & Proposed & \textbf{76.4} & \textbf{76.3} & \textbf{76.6} \\ \midrule
\multirow{3}{*}{FOOD101} & w/o $\mathcal{L}_{\mathrm{o}}$ & 73.4 & 73.3 & 73.3 \\ 
                  & w/o $\mathcal{L}_{\mathrm{a}}$ & 61.7 & 61.4 & 61.5 \\ 
                  & Proposed & \textbf{77.5} & \textbf{77.6} & \textbf{77.6} \\ \midrule
            Dataset   & Method  & ACC & F1 & AUC \\ \midrule
\multirow{3}{*}{ROSMAP} & w/o $\mathcal{L}_{\mathrm{o}}$ & 77.4 & 77.2 & 82.7 \\ 
                  & w/o $\mathcal{L}_{\mathrm{a}}$ & 55.2 & 55.4 & 57.1 \\
                  & Proposed & \textbf{79.2} & \textbf{79.6} & \textbf{84.5} \\ \bottomrule
\end{tabular}
}
\label{tab:constraint-cm}
    \end{subtable}
    \setlength{\tabcolsep}{6pt}
  \end{minipage}

\end{table}

\subsubsection{Effectiveness of $\mathcal{L}_{\mathrm{o}}$ and $\mathcal{L}_{\mathrm{a}}$}
Further experiments are conducted to evaluate the roles of the two constraints \(\mathcal{L}_{\mathrm{o}}\) and \(\mathcal{L}_{\mathrm{a}}\) in noise removal. We perform ablation studies on datasets corrupted only by modality-specific noise with \(\epsilon = 5\) and only by cross-modality noise with \(\eta = 10\%\), respectively. As shown in Table \ref{tab:contraint-ms}, on datasets with only modality-specific noise \((\epsilon = 5)\), removing \(\mathcal{L}_{\mathrm{o}}\) causes a much more significant performance drop than removing \(\mathcal{L}_{\mathrm{a}}\). This indicates that \(\mathcal{L}_{\mathrm{o}}\) plays a dominant role in guiding modality-specific noise removal. In contrast, as shown in Table \ref{tab:constraint-cm}, on datasets with only cross-modality noise \((\eta = 10\%)\), removing \(\mathcal{L}_{\mathrm{a}}\) results in a substantially larger performance drop compared to removing \(\mathcal{L}_{\mathrm{o}}\), demonstrating that \(\mathcal{L}_{\mathrm{a}}\) primarily guides cross-modality noise removal.

\begin{table}[t]
\caption{Accuracy of alignment on projection and without projection under cross-modality noise with varying \(\eta\) (with modality-specific noise \(\epsilon=5\)).}
\begin{center}
\renewcommand{\arraystretch}{0.6}
\resizebox{0.6\linewidth}{!}{
\begin{tabular}{ccccc}
\toprule
Dataset               & Method & $\eta=0$ & $\eta=10\%$ & $\eta=20\%$\\ \midrule
\multirow{2}{*}{BRCA}   & w/o projection     & 84.1 & 60.2 & 51.8  \\
    & with projection      & \textbf{84.4} & \textbf{73.2} & \textbf{72.8}\\ 
\midrule

\multirow{2}{*}{ROSMAP}  & w/o projection     & 80.7 & 71.2 & 58.3  \\
     & with projection      & \textbf{81.1} & \textbf{76.4} & \textbf{71.8} \\ 
\midrule
\multirow{2}{*}{CUB} & w/o projection     & 93.1 & 67.8 & 57.9  \\
    & with projection    & \textbf{93.6} & \textbf{75.3}& \textbf{70.2} \\ 
                      \midrule

\multirow{2}{*}{FOOD101}  & w/o projection     & 93.8 & 69.2 & 58.6  \\
     & with projection     & \textbf{94.0} & \textbf{76.3} & \textbf{70.4}  \\ 
                      \bottomrule
\end{tabular}
}
\end{center}
\label{tab:sp}
\end{table}

\subsubsection{Further Discussion On the Subspace Projection Alignment} To further verify the effectiveness of Subspace Projection Alignment in aligning projections onto stable feature subspaces, we add cross-modality noise with varying intensity \(\eta\) on data with modality-specific noise \(\epsilon=5\). We compare two strategies: aligning directly on modality features (w/o projection) and aligning projections onto stable subspaces via Subspace Projection Alignment (with projection). ``w/o projection'' is achieved by directly minimizing the Frobenius norm of differences between modality features. As shown in Table \ref{tab:sp}, under modality-specific noise, the ``with projection'' strategy exhibits stronger robustness than ``w/o projection'' as cross-modality noise increases. This indicates that aligning modality projections within their respective stable subspaces enables more effective suppression of cross-modality noise in the presence of modality-specific noise.

\begin{figure}[t]
\centering
\includegraphics[width=\linewidth]{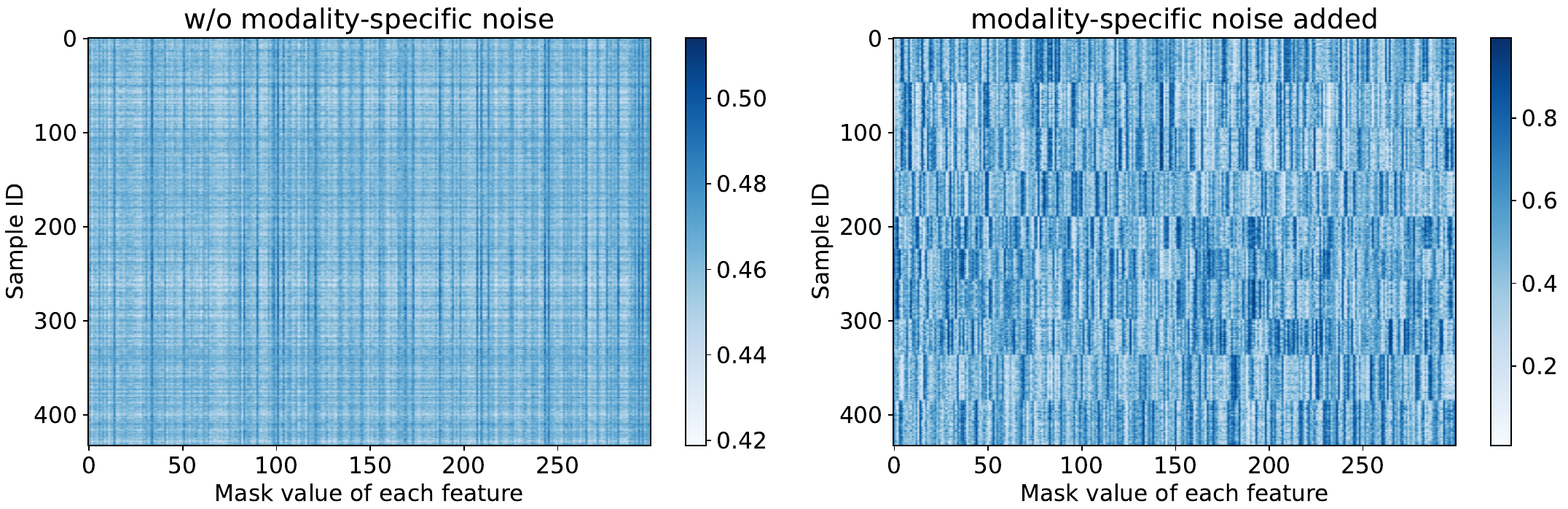} 
\caption{Visualization of sample-wise masks learned by the modality-specific noise expert on the corrupted image modality of CUB, before and after adding modality-specific noise (\(\epsilon=5\)).}
\label{fig:ms-mask}
\end{figure}

\begin{figure}[t]
\centering
\includegraphics[width=\linewidth]{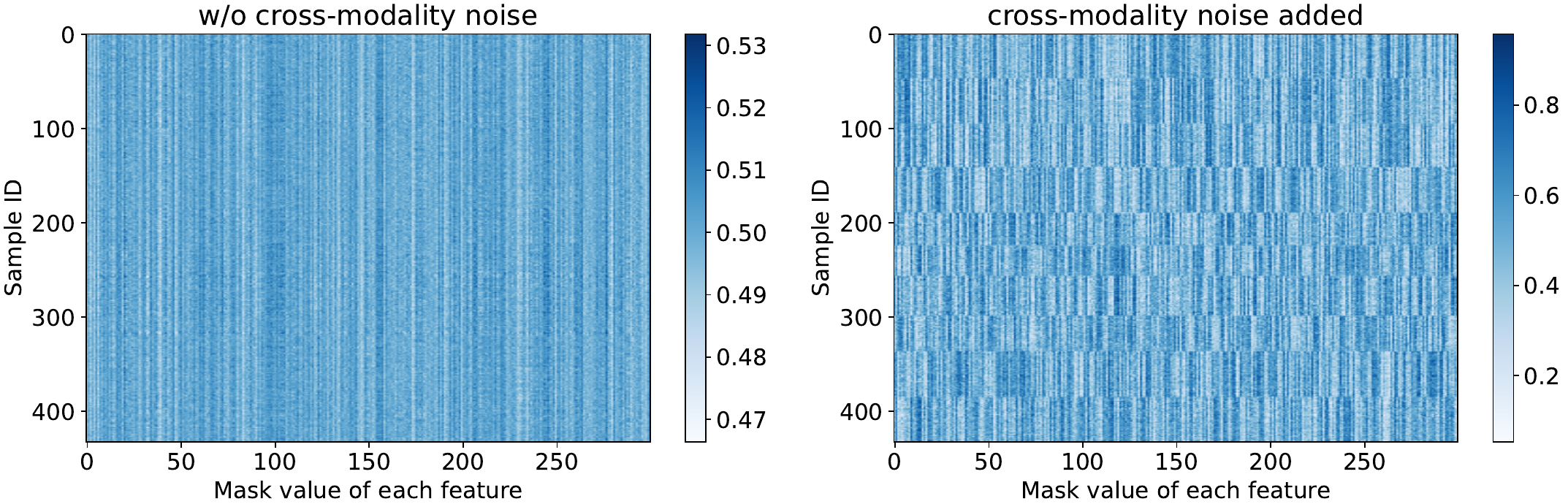} 
\caption{Visualization of sample-wise masks learned by the cross-modality noise expert on the corrupted image modality of CUB, before and after adding cross-modality noise (\(\eta=10\%\)).}
\label{fig:cm-mask}
\end{figure}

\subsection{Discussion On SACA} \label{sec:dis-smne}
\subsubsection{Effectiveness of Noise Experts}
More experiments are conducted to verify that the modality specific noise experts and cross modality noise experts in SACA respectively facilitate the removal of instance level modality-specific and cross-modality noise. Specifically, we first introduce only modality-specific noise $\epsilon=5$ into one modality of the dataset and visualize the sample masks $\{w_{s,i}^m\}_{i=1}^N$ learned by the corresponding modality-specific noise expert. Figure \ref{fig:ms-mask} illustrates the masks learned by the modality-specific noise expert on the image modality affected by noise in the CUB dataset. In the noise-free setting (left), the mask values exhibit relatively strong consistency across samples. After modality-specific noise is added (right), the learned mask shows increased variability between samples, showing that the modality-specific noise expert effectively captures the instance-level chaos introduced by the modality noise. Also, we introduce cross-modality noise with \(\eta = 20\%\) to one modality of the dataset by randomly shuffling the features of 20\% of its samples. Figure \ref{fig:cm-mask} shows the masks learned by the cross-modality noise expert on the image modality affected by noise in the CUB dataset. Before noise is added (left), the masks of different samples exhibit a relatively consistent pattern, whereas after noise introduction (right), the masks show substantial variability across samples. This demonstrates that the cross-modality noise expert can also effectively capture sample-specific corrupted information introduced by noise.



\begin{table}[t]
\caption{Comparison of model accuracy under noise using different cross-modality alignment methods.}
\begin{center}
\renewcommand{\arraystretch}{0.6}
\resizebox{0.6\linewidth}{!}{
\begin{tabular}{ccccc}
\toprule
Dataset               & Method & $\eta=0$ & $\eta=10\%$ & $\eta=20\%$\\ \midrule
\multirow{3}{*}{BRCA}   & \text{sim}     & 84.6 & 66.8 & 57.3  \\
    & \text{MI}     & 85.0 & 70.8 & 61.3  \\
    & slack      & \textbf{85.6} & \textbf{77.4} & \textbf{72.8}\\ 
\midrule

\multirow{3}{*}{ROSMAP}  & \text{sim}     & 86.2 & 75.3 & 62.8  \\
& \text{MI}     & 86.9 & 76.4 & 65.1  \\
     & slack      & \textbf{87.7} & \textbf{79.2} & \textbf{71.8} \\ 
\midrule
\multirow{3}{*}{CUB} & \text{sim}     & 92.0 & 69.8 & 60.5  \\
& \text{MI}     & 92.3 & 70.6 & 62.7  \\
    & slack    & \textbf{93.8} & \textbf{76.4}& \textbf{70.2} \\ 
                      \midrule

\multirow{3}{*}{FOOD101}  & \text{sim}     & 92.5 & 71.3 & 61.8  \\
& \text{MI}     & 93.0 & 72.6 & 64.1  \\
     & slack     & \textbf{94.2} & \textbf{77.5} & \textbf{70.4}  \\ 
                      \bottomrule
\end{tabular}
}
\end{center}
\label{tab:differnt-align}
\end{table}

\subsubsection{Discussion on Confidence-Aware Asymmetric Slack Alignment}
We conduct experiments to validate the effectiveness of the slack alignment proposed in Confidence-Aware Asymmetric Slack Alignment. Specifically, we compare with existing methods that enforce cross-modality consistency, including similarity-based approaches (sim), as used in \cite{xie2025seeking,dang2025disentangled}, and information-theoretic approaches (MI), as used in \cite{zeng2023semantic}. ``slack'' represents applying our $\mathcal{L}_{\mathrm{nll}}^c$ that retains the slack-alignment mechanism while removing modality-confidence guidance. As shown in Table \ref{tab:differnt-align}, under cross-modality noise with varying intensity \(\eta\), our slack alignment exhibits stronger stability than other methods, indicating superior robustness. ``MI'' achieves better performance than ``sim'', likely due to its less restrictive constraint on cross-modality consistency, preserving more complementary information across modalities.

\begin{figure}[t]
\centering
\includegraphics[width=\linewidth]{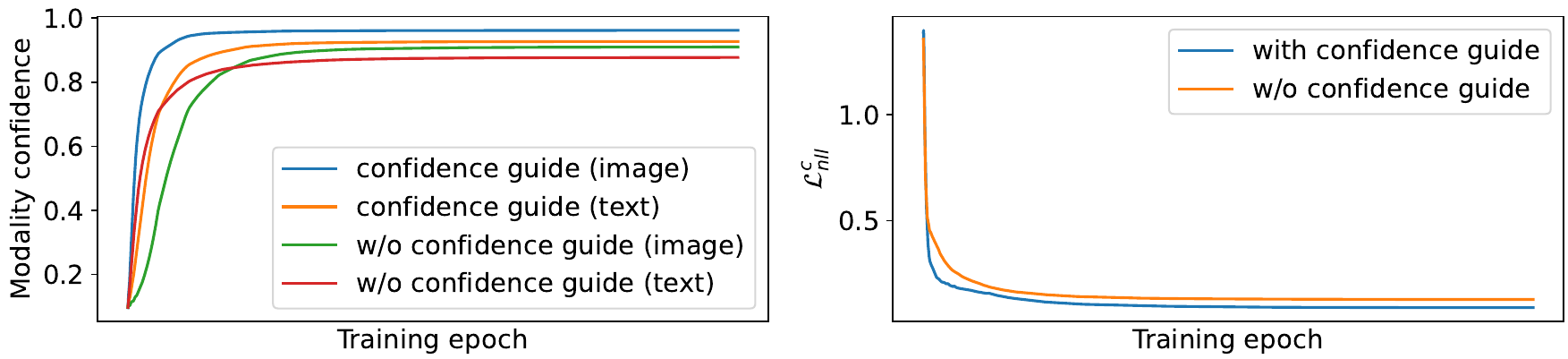} 
\caption{Changes in confidence of each modality and the decline of \(\mathcal{L}_{\mathrm{nll}}^c\) during training on the FOOD101 dataset with cross-modality noise \((\eta = 10\%)\) and modality-specific noise \((\epsilon=1)\), under two conditions: with and without the guidance of modality confidence on slack alignment.
}
\label{fig:conf-aware}
\end{figure}

We also conduct experiment to validate the effectiveness of incorporating modality confidence to guide slack alignment. Specifically, the model is trained on data with modality-specific and cross-modality noise ($\epsilon=1,\eta=10\%$), and the changes in modality-wise confidence and the loss \(\mathcal{L}_{\mathrm{nll}}^c\) are reported for both with and without confidence guidance. The variant without confidence guidance is implemented by removing the gradient modulation in Eq. (\ref{eq:aym-conf}). Modality confidence reported in Figure \ref{fig:conf-aware} is computed as the average of the likelihoods \(p^m(\hat{h}_{s,i}^m)\) of all sample representations \(\hat{h}_{s,i}^m\) under the corresponding modality-specific feature prior \(p^m\). As shown in the left panel of Figure \ref{fig:conf-aware}, as training progresses, the confidence of each modality gradually increases, indicating progressive noise removal. Moreover, with modality-confidence guidance, modality confidence rises to higher levels than without such guidance, suggesting that confidence guidance promotes alignment toward high-confidence features across modalities, thereby yielding more robust representations. Moreover, as shown in the right panel of Figure \ref{fig:conf-aware}, the loss \(\mathcal{L}_{\mathrm{nll}}^c\) decreases faster with confidence guidance than without it, indicating that confidence guidance also accelerates the alignment process.

\begin{figure}[t]
    \centering
    \begin{subfigure}[b]{0.47\linewidth}
        \includegraphics[width=\linewidth]{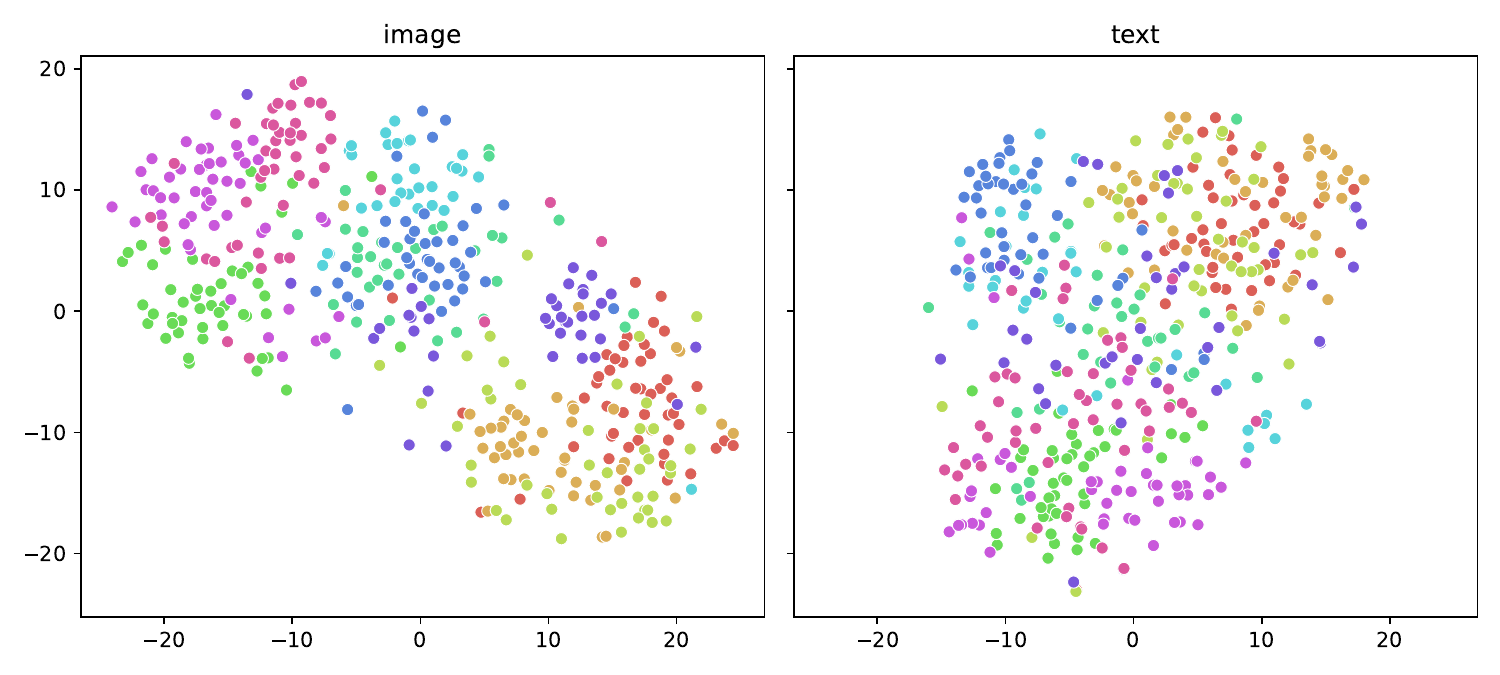}
        \caption{0 Iteration ($\mathcal{L}_{\mathrm{nll}}^c=1.3647$)}
        \label{fig:iter0}
    \end{subfigure}
    \hfill
    \begin{subfigure}[b]{0.47\linewidth}
        \includegraphics[width=\linewidth]{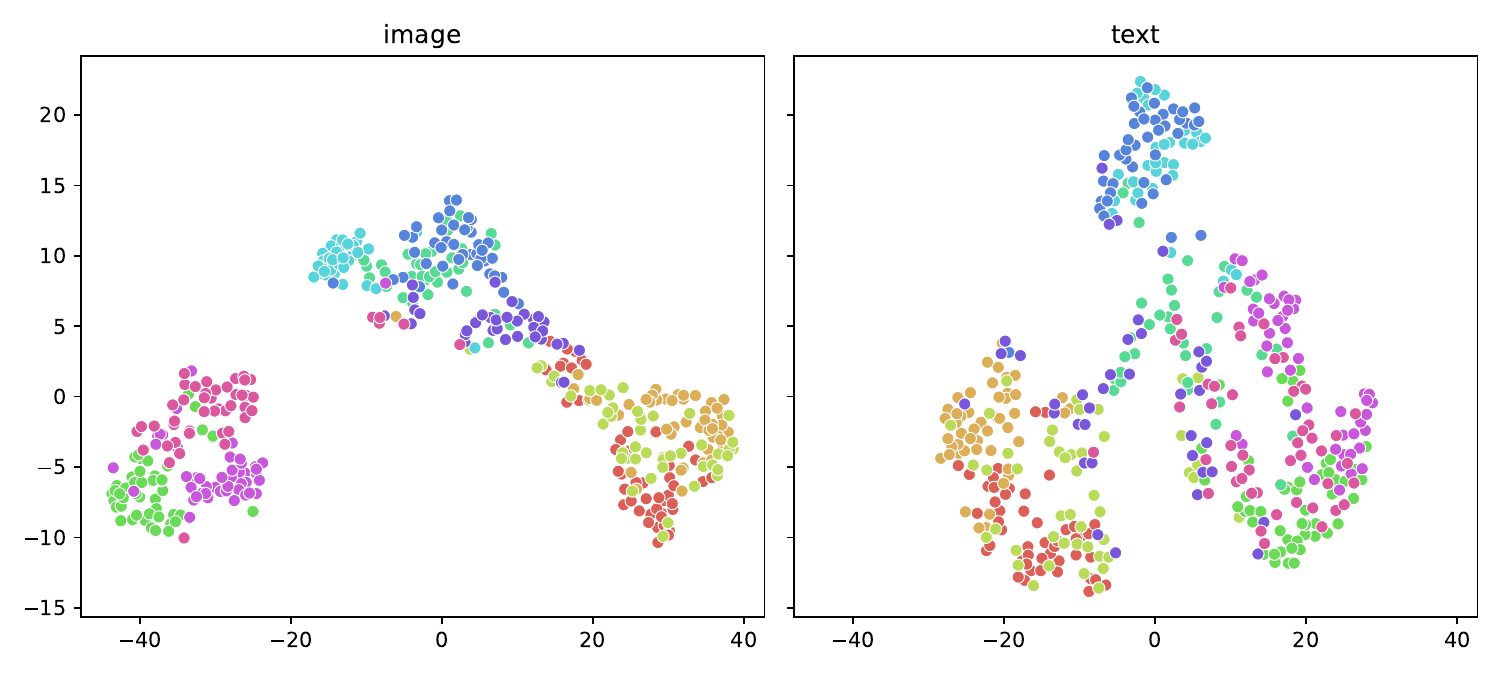}
        \caption{10 Iteration ($\mathcal{L}_{\mathrm{nll}}^c=0.6608$)}
        \label{fig:iter70}
    \end{subfigure}

    \vskip\baselineskip
    \begin{subfigure}[b]{0.47\linewidth}
        \includegraphics[width=\linewidth]{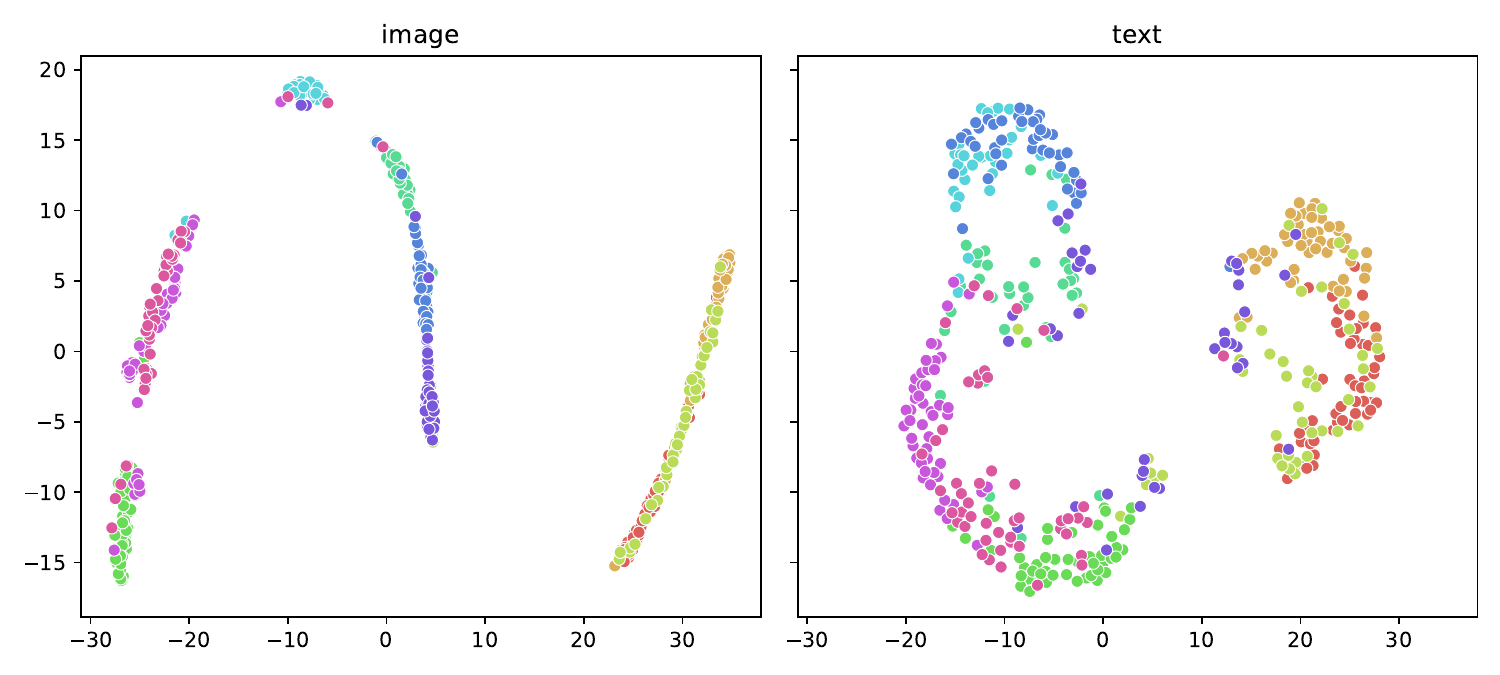}
        \caption{20 Iteration ($\mathcal{L}_{\mathrm{nll}}^c=0.3561$)}
        \label{fig:iter270}
    \end{subfigure}
    \hfill
    \begin{subfigure}[b]{0.47\linewidth}
        \includegraphics[width=\linewidth]{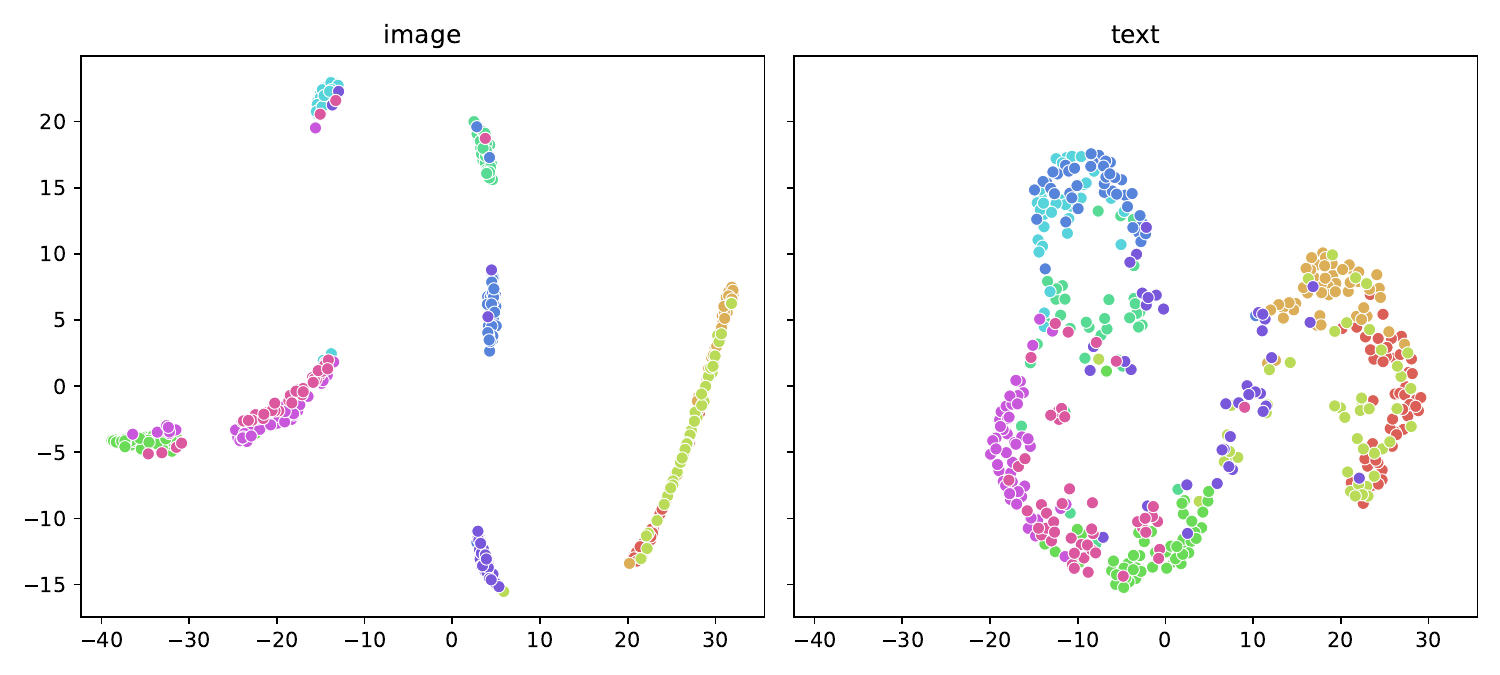}
        \caption{30 Iteration ($\mathcal{L}_{\mathrm{nll}}^c=0.2910$)}
        \label{fig:iter300}
    \end{subfigure}
    \caption{T-SNE visualization of the learned modality representations and the value of cross-modality alignment loss \(\mathcal{L}_{\mathrm{nll}}^c\) results on the CUB dataset with both modalities corrupted by modality-specific and cross-modality noise (\(\epsilon=5, \eta=10\%\)) after different TTCE co-enhancement iterations. Points in different colors represent sample representations from different classes.}
    \label{fig:ttce}
\end{figure}

\subsection{Discussion On TTCE}\label{sec:dis-ttce}
\subsubsection{Adaptation And Enhancement against Unseen Noise}
To verify that TTCE effectively enhances the model’s adaptability to previously unseen noise, we conduct experiments using the original training data and test data corrupted with modality-specific and cross-modality noise ($\eta=10\%,\epsilon=5$). The model is first trained, and then the representations $\hat{h}_s^m+\hat{h}_c^m$ of each modality during the iterative enhancement of TTCE on the noisy test data are visualized. As shown in Figure \ref{fig:ttce}, before TTCE iterations (0 iteration), the learned representations exhibit poor intra-modality class separability and weak cross-modality alignment (high $\mathcal{L}_{\mathrm{nll}}^c$), indicating insufficient removal of newly encountered noise. As TTCE proceeds, intra-modality class separability gradually improves, and cross-modality alignment is substantially enhanced ($\mathcal{L}_{\mathrm{nll}}^c$ decreases), indicating that newly introduced modality-specific and cross-modality noise is progressively removed. These results demonstrate that, through multiple TTCE updates, the model can progressively adapt to and remove noise unseen in training data, thereby achieving stronger generalization performance.

\begin{figure}[t]
\centering
\includegraphics[width=\columnwidth]{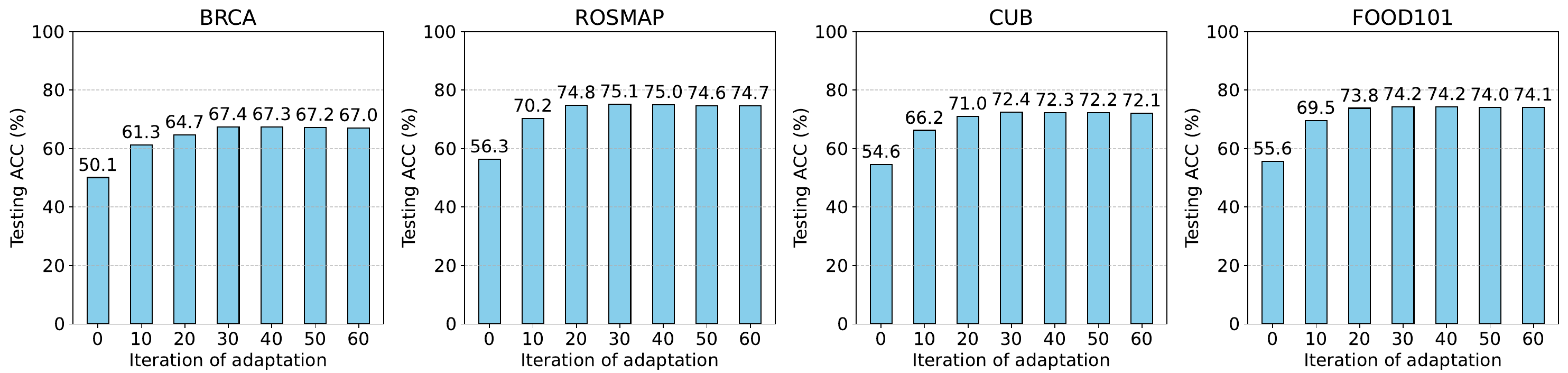} 
\caption{Analysis of TTCE iteration number \(E\) on four datasets.}
\label{fig:inter-adapt}
\end{figure}

\begin{table}[t]
\caption{Comparison of generalization performance and runtime across methods on the BRCA dataset. Cross-modality noise and modality-specific noise with severities \(\eta\) and \(\epsilon\) are added only onto the test data. The runtime for TAHCD is reported with the best TTCE iteration \(E\) under each noise condition shown in parentheses.}
\begin{center}
\renewcommand{\arraystretch}{0.6}
\resizebox{0.6\linewidth}{!}{
\begin{tabular}{c|c|cccc}
\toprule
Data Type               & Accuracy and Efficiency & MD & PDF & SPS & TAHCD\\ \midrule
\multirow{2}{*}{\begin{tabular}[c]{@{}c@{}}$\eta=0$\\ $\epsilon=0$ \end{tabular}}   & ACC (\%)     & 82.1 & 83.7 & 79.8 & \textbf{85.6}  \\
    & Runtime (ms)      & 160.22 & 124.67 & 52.03 & 99.23 ($E=5$)\\ 
\midrule

\multirow{2}{*}{\begin{tabular}[c]{@{}c@{}}$\eta=5\%$\\ $\epsilon=1$ \end{tabular}}  & ACC (\%)     & 69.6 & 72.1 & 66.3 & \textbf{79.5}  \\
     & Runtime (ms)      & 160.22 & 124.67 & 52.03 & 136.35 ($E=20$) \\ 
\midrule
\multirow{2}{*}{\begin{tabular}[c]{@{}c@{}}$\eta=10\%$\\ $\epsilon=5$ \end{tabular}} & ACC (\%)     & 51.4 & 56.6 & 46.1 & \textbf{67.4}  \\
    & Runtime (ms)    & 160.22 & 124.67& 52.03 & 165.28 ($E=30$) \\ 
\bottomrule
\end{tabular}
}
\end{center}
\label{tab:acc_and_eff}
\end{table}

\begin{table}[t]
\caption{Top 5 informative biomarkers that TAHCD learned from different modalities on BRCA dataset.}
\begin{center}
\renewcommand{\arraystretch}{0.6}
\resizebox{0.6\linewidth}{!}{
\begin{tabular}{c|c}
\hline
Modality & Top 5 informative biomarkers\\ \hline
DNA & ZNF671, KRTAP3-1, AGR2, MYT1, RPS6KL1 \\ \hline
mRNA & KLK8, PTX3, SOX11, KRT6B, RDH10 \\ \hline
miRNA & mir-21, mir-190b, mir-378, mir-93, mir-23b \\
\hline
\end{tabular}}
\end{center}
\label{tab:bi}
\end{table}

\subsubsection{Computational Efficiency}
Additional experiments are conducted to evaluate the computational efficiency of TTCE under noisy conditions. Since TTCE does not introduce additional network structures or parameters, we focus on evaluating runtime of the model during testing. As shown in Table \ref{tab:acc_and_eff}, modality-specific and cross-modality noise with varying severities \(\epsilon\) and \(\eta\) are added into the two modalities of the CUB testing data, and we compare the test accuracy and runtime of different methods. We evaluate the test accuracy of TAHCD every five enhancement iterations in the experiment. It can be observed that TAHCD achieves superior and more stable performance while maintaining competitive runtime, demonstrating its high computational efficiency. Under noise-free condition ($\eta=0,\epsilon=0$), TAHCD exhibits lower runtime than MD and PDF. As noise increases, TTCE in TAHCD requires more enhancement iterations $E$, resulting in longer runtime. Even so, under \(\eta = 10\%\) and \(\epsilon = 5\), TAHCD maintains runtime comparable to MD and PDF while achieving over 10\% higher accuracy.

\subsubsection{Discussion on the Number of Enhancement Iterations \(E\)} \label{sec:dis-E}
We add modality-specific noise with \(\epsilon = 5\) and cross-modality noise with \(\eta = 10\%\) into the test sets of four datasets and report the test accuracy of TAHCD under different values of \(E\). Figure \ref{fig:inter-adapt} shows that as the number of update iterations \(E\) increases, the test accuracy first rises and generally stabilizes around \(E=30\) across all datasets. This indicates that the model has completed its adaptive adjustment to the noise present in the current test data. Moreover, Table \ref{tab:acc_and_eff} shows that when the noise severity increases, the optimal value of \(E\) corresponding to the highest test accuracy increases, indicating that more enhancement iterations are required.

\subsection{Biomarkers Identification}\label{sec:bi}
To validate that TAHCD reliably learns informative features, following previous works \cite{han2022multimodal,zou2023dpnet}, we examine the biomarkers identified by TAHCD for early diagnosis and prognosis on the multi-omic BRCA dataset. Specifically, we compute each modality’s feature informativeness as the mean of the product between the global-level informativeness \(w_{\lambda}^m\) learned in Eq. (\ref{eq:adaptive-filter}) and the instance-level informativeness \(w_{s,i}^m\) learned in Eq. (\ref{eq:exp-msn-remove}) by TAHCD. The results in Table \ref{tab:bi} show that the biomarkers identified by TAHCD are supported by existing medical studies as key indicators for cancer diagnosis and prognosis. For instance, KLK8 has been reported as an independent prognostic indicator in breast cancer patients \cite{michaelidou2015clinical}, while low expression of ZNF671 \cite{zhang2019epigenetic} and miR-378 \cite{arabkari2023mirna} are both linked to poor prognosis in breast cancer.

\begin{figure}[t]
\centering
\includegraphics[width=\linewidth]{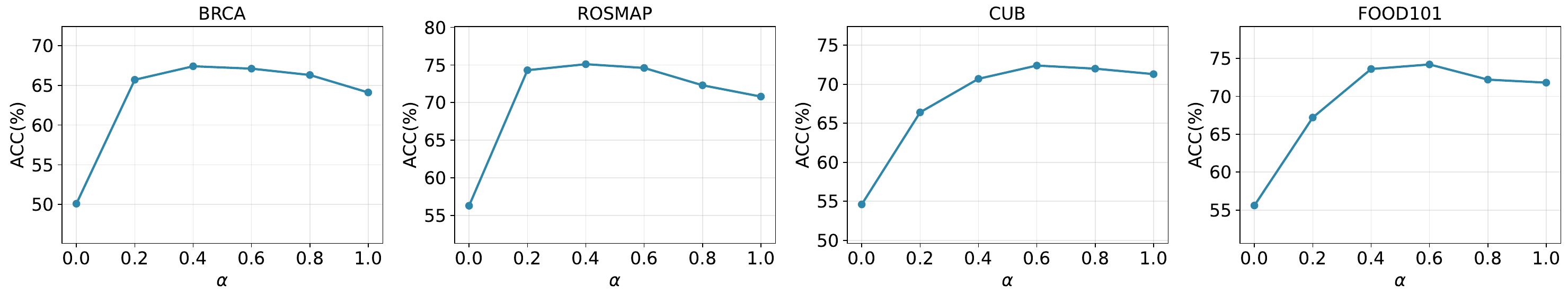} 
\caption{Variation of test accuracy under different values of \(\alpha\).}
\label{fig:param-analysis}
\end{figure}

\subsection{Parameter Analysis}\label{sec:pa}
More experiments are conducted to analyze the hyperparameter $\alpha$ of TAHCD defined in Eq. (\ref{eq:update-mu-Sigma}). $\alpha\in[0,1]$ indicates the extent to which the modality feature distributions incorporate the newly estimated mean and covariance, where $\alpha=0$ means that the distributions are not updated, and $\alpha=1$ indicates that the newly computed statistics fully replace the original ones. Figure \ref{fig:param-analysis} shows that as \(\alpha\) increases, the accuracy of TAHCD first rises and then declines. The model achieves substantially better performance when \(\alpha \neq 0\) than when \(\alpha = 0\), demonstrating the effectiveness of TAHCD’s adaptive updates in improving generalization to unseen noise. Moreover, TAHCD achieves high accuracy across a relatively wide range of \(\alpha\), indicating low sensitivity to this hyperparameter.


\section{Conclusion}
Existing multimodal learning methods struggle to achieve robust multimodal representation learning under heterogeneous noise and exhibit limited adaptability and generalization to unseen noise. To address these challenges, we propose the Test-time Adaptive Hierarchical Co-enhanced Denoising Network (TAHCD). TAHCD introduces Adaptive Stable Subspace Alignment and Sample-Adaptive Confidence Alignment to jointly handle modality-specific and cross-modality noise at the global and instance levels, respectively, thereby improving robustness. In addition, TAHCD designs a Test-Time Cooperative Enhancement mechanism that enables cooperative enhancement between global-level and instance-level denoising processes driven by input sample noise in a label-free manner. This enables TAHCD to adapt more effectively to unseen noise at test time, improving the model's generalization ability. Experiments on multiple benchmarks demonstrate that TAHCD outperforms other state-of-the-art reliable multimodal learning methods in robustness and generalization ability.

Although existing reliable multimodal learning methods have been extensively studied for classification tasks, their reliability in other machine learning tasks, such as regression, remains insufficiently explored. This presents a promising and important direction for further investigation. Accordingly, future work will extend and validate the proposed approach across a broader set of machine learning tasks.

\bibliographystyle{elsarticle-num-names}
\bibliography{reference.bib}

@inproceedings{radenovic2023filtering,
  title={Filtering, distillation, and hard negatives for vision-language pre-training},
  author={Radenovic, Filip and Dubey, Abhimanyu and Kadian, Abhishek and Mihaylov, Todor and Vandenhende, Simon and Patel, Yash and Wen, Yi and Ramanathan, Vignesh and Mahajan, Dhruv},
  booktitle={Proceedings of the IEEE/CVF conference on computer vision and pattern recognition},
  pages={6967--6977},
  year={2023}
}

@inproceedings{geng2021uncertainty,
  title={Uncertainty-aware multi-view representation learning},
  author={Geng, Yu and Han, Zongbo and Zhang, Changqing and Hu, Qinghua},
  booktitle={Proceedings of the AAAI Conference on Artificial Intelligence},
  volume={35},
  number={9},
  pages={7545--7553},
  year={2021}
}

@inproceedings{han2022multimodal,
  title={Multimodal dynamics: Dynamical fusion for trustworthy multimodal classification},
  author={Han, Zongbo and Yang, Fan and Huang, Junzhou and Zhang, Changqing and Yao, Jianhua},
  booktitle={Proceedings of the IEEE/CVF conference on computer vision and pattern recognition},
  pages={20707--20717},
  year={2022}
}

@inproceedings{zhang2023provable,
  title={Provable dynamic fusion for low-quality multimodal data},
  author={Zhang, Qingyang and Wu, Haitao and Zhang, Changqing and Hu, Qinghua and Fu, Huazhu and Zhou, Joey Tianyi and Peng, Xi},
  booktitle={International conference on machine learning},
  pages={41753--41769},
  year={2023},
  organization={PMLR}
}

@inproceedings{zheng2023multi,
  title={Multi-level confidence learning for trustworthy multimodal classification},
  author={Zheng, Xiao and Tang, Chang and Wan, Zhiguo and Hu, Chengyu and Zhang, Wei},
  booktitle={Proceedings of the AAAI Conference on Artificial Intelligence},
  volume={37},
  number={9},
  pages={11381--11389},
  year={2023}
}

@inproceedings{zou2023dpnet,
  title={DPNET: Dynamic Poly-attention Network for Trustworthy Multi-modal Classification},
  author={Zou, Xin and Tang, Chang and Zheng, Xiao and Li, Zhenglai and He, Xiao and An, Shan and Liu, Xinwang},
  booktitle={Proceedings of the 31st ACM International Conference on Multimedia},
  pages={3550--3559},
  year={2023}
}

@inproceedings{zhou2023calm,
  title={CALM: An Enhanced Encoding and Confidence Evaluating Framework for Trustworthy Multi-view Learning},
  author={Zhou, Hai and Xue, Zhe and Liu, Ying and Li, Boang and Du, Junping and Liang, Meiyu and Qi, Yuankai},
  booktitle={Proceedings of the 31st ACM International Conference on Multimedia},
  pages={3108--3116},
  year={2023}
}

@inproceedings{
cao2024predictive,
title={Predictive Dynamic Fusion},
author={Bing Cao and Yinan Xia and Yi Ding and Changqing Zhang and Qinghua Hu},
booktitle={Forty-first International Conference on Machine Learning},
year={2024},
url={https://openreview.net/forum?id=LYpGLrC4oq}
}

@article{zheng2024global,
  title={Global and cross-modal feature aggregation for multi-omics data classification and application on drug response prediction},
  author={Zheng, Xiao and Wang, Minhui and Huang, Kai and Zhu, En},
  journal={Information Fusion},
  volume={102},
  pages={102077},
  year={2024},
  publisher={Elsevier}
}

@article{zhang2024multimodal,
  title={Multimodal fusion on low-quality data: A comprehensive survey},
  author={Zhang, Qingyang and Wei, Yake and Han, Zongbo and Fu, Huazhu and Peng, Xi and Deng, Cheng and Hu, Qinghua and Xu, Cai and Wen, Jie and Hu, Di and others},
  journal={arXiv preprint arXiv:2404.18947},
  year={2024}
}

@inproceedings{changpinyo2021conceptual,
  title={Conceptual 12m: Pushing web-scale image-text pre-training to recognize long-tail visual concepts},
  author={Changpinyo, Soravit and Sharma, Piyush and Ding, Nan and Soricut, Radu},
  booktitle={Proceedings of the IEEE/CVF conference on computer vision and pattern recognition},
  pages={3558--3568},
  year={2021}
}

@article{huang2021learning,
  title={Learning with noisy correspondence for cross-modal matching},
  author={Huang, Zhenyu and Niu, Guocheng and Liu, Xiao and Ding, Wenbiao and Xiao, Xinyan and Wu, Hua and Peng, Xi},
  journal={Advances in Neural Information Processing Systems},
  volume={34},
  pages={29406--29419},
  year={2021}
}

@inproceedings{huang2023nlip,
  title={Nlip: Noise-robust language-image pre-training},
  author={Huang, Runhui and Long, Yanxin and Han, Jianhua and Xu, Hang and Liang, Xiwen and Xu, Chunjing and Liang, Xiaodan},
  booktitle={Proceedings of the AAAI Conference on Artificial Intelligence},
  volume={37},
  number={1},
  pages={926--934},
  year={2023}
}

@inproceedings{nakada2023understanding,
  title={Understanding multimodal contrastive learning and incorporating unpaired data},
  author={Nakada, Ryumei and Gulluk, Halil Ibrahim and Deng, Zhun and Ji, Wenlong and Zou, James and Zhang, Linjun},
  booktitle={International Conference on Artificial Intelligence and Statistics},
  pages={4348--4380},
  year={2023},
  organization={PMLR}
}

@article{zeng2023semantic,
  title={Semantic invariant multi-view clustering with fully incomplete information},
  author={Zeng, Pengxin and Yang, Mouxing and Lu, Yiding and Zhang, Changqing and Hu, Peng and Peng, Xi},
  journal={IEEE Transactions on Pattern Analysis and Machine Intelligence},
  year={2023},
  publisher={IEEE}
}

@article{han2022trusted,
  title={Trusted multi-view classification with dynamic evidential fusion},
  author={Han, Zongbo and Zhang, Changqing and Fu, Huazhu and Zhou, Joey Tianyi},
  journal={IEEE transactions on pattern analysis and machine intelligence},
  volume={45},
  number={2},
  pages={2551--2566},
  year={2022},
  publisher={IEEE}
}

@article{a2012overview,
  title={Overview and findings from the religious orders study},
  author={A Bennett, David and A Schneider, Julie and Arvanitakis, Zoe and S Wilson, Robert},
  journal={Current Alzheimer Research},
  volume={9},
  number={6},
  pages={628--645},
  year={2012},
  publisher={Bentham Science Publishers}
}

@misc{de2018multi,
  title={A multi-omic atlas of the human frontal cortex for aging and Alzheimer’s disease research. Sci Data 5: 180142},
  author={De Jager, PL and Ma, Y and McCabe, C and Xu, J and Vardarajan, BN and Felsky, D and Klein, HU and White, CC and Peters, MA and Lodgson, B and others},
  year={2018}
}

@article{wah2011caltech,
  title={The caltech-ucsd birds-200-2011 dataset},
  author={Wah, Catherine and Branson, Steve and Welinder, Peter and Perona, Pietro and Belongie, Serge},
  year={2011},
  publisher={California Institute of Technology}
}

@inproceedings{wang2015recipe,
  title={Recipe recognition with large multimodal food dataset},
  author={Wang, Xin and Kumar, Devinder and Thome, Nicolas and Cord, Matthieu and Precioso, Frederic},
  booktitle={2015 IEEE International Conference on Multimedia \& Expo Workshops (ICMEW)},
  pages={1--6},
  year={2015},
  organization={IEEE}
}

@misc{lingle9cancer,
  title={The cancer genome atlas breast invasive carcinoma collection (TCGA-BRCA)(Version 3)[Data set]. Cancer Imag. Arch.(2016)},
  author={Lingle, W and others}
}

@article{mukherjee2015religious,
  title={Religious Orders Study/Memory and Aging Project Investigators; Alzheimer’s Disease Genetics Consortium},
  author={Mukherjee, S and Walter, S and Kauwe, JSK and Adult Changes in Thought Study Investigators and others},
  journal={Genetically predicted body mass index and Alzheimer’s disease-related phenotypes in three large samples: Mendelian randomization analyses. Alzheimers Dement},
  volume={11},
  number={12},
  pages={1439--1451},
  year={2015}
}

@article{baltruvsaitis2018multimodal,
  title={Multimodal machine learning: A survey and taxonomy},
  author={Baltru{\v{s}}aitis, Tadas and Ahuja, Chaitanya and Morency, Louis-Philippe},
  journal={IEEE transactions on pattern analysis and machine intelligence},
  volume={41},
  number={2},
  pages={423--443},
  year={2018},
  publisher={IEEE}
}

@inproceedings{lee2021variational,
  title={A variational information bottleneck approach to multi-omics data integration},
  author={Lee, Changhee and Van der Schaar, Mihaela},
  booktitle={International Conference on Artificial Intelligence and Statistics},
  pages={1513--1521},
  year={2021},
  organization={PMLR}
}

@article{wang2021mogonet,
  title={MOGONET integrates multi-omics data using graph convolutional networks allowing patient classification and biomarker identification},
  author={Wang, Tongxin and Shao, Wei and Huang, Zhi and Tang, Haixu and Zhang, Jie and Ding, Zhengming and Huang, Kun},
  journal={Nature communications},
  volume={12},
  number={1},
  pages={3445},
  year={2021},
  publisher={Nature Publishing Group UK London}
}

@article{gadre2024datacomp,
  title={Datacomp: In search of the next generation of multimodal datasets},
  author={Gadre, Samir Yitzhak and Ilharco, Gabriel and Fang, Alex and Hayase, Jonathan and Smyrnis, Georgios and Nguyen, Thao and Marten, Ryan and Wortsman, Mitchell and Ghosh, Dhruba and Zhang, Jieyu and others},
  journal={Advances in Neural Information Processing Systems},
  volume={36},
  year={2024}
}

@ARTICLE{9906429,
  author={Wang, Yikai and Sun, Fuchun and Huang, Wenbing and He, Fengxiang and Tao, Dacheng},
  journal={IEEE Transactions on Pattern Analysis and Machine Intelligence}, 
  title={Channel Exchanging Networks for Multimodal and Multitask Dense Image Prediction}, 
  year={2023},
  volume={45},
  number={5},
  pages={5481-5496},
  doi={10.1109/TPAMI.2022.3211086}}

@article{zhu2024vision+,
  title={Vision+ x: A survey on multimodal learning in the light of data},
  author={Zhu, Ye and Wu, Yu and Sebe, Nicu and Yan, Yan},
  journal={IEEE Transactions on Pattern Analysis and Machine Intelligence},
  year={2024},
  publisher={IEEE}
}

@article{liang2022foundations,
  title={Foundations and trends in multimodal machine learning: Principles, challenges, and open questions},
  author={Liang, Paul Pu and Zadeh, Amir and Morency, Louis-Philippe},
  journal={arXiv preprint arXiv:2209.03430},
  year={2022}
}

@InProceedings{pmlr-v235-cao24c,
  title = 	 {Predictive Dynamic Fusion},
  author =       {Cao, Bing and Xia, Yinan and Ding, Yi and Zhang, Changqing and Hu, Qinghua},
  booktitle = 	 {Proceedings of the 41st International Conference on Machine Learning},
  pages = 	 {5608--5628},
  year = 	 {2024},
  editor = 	 {Salakhutdinov, Ruslan and Kolter, Zico and Heller, Katherine and Weller, Adrian and Oliver, Nuria and Scarlett, Jonathan and Berkenkamp, Felix},
  volume = 	 {235},
  series = 	 {Proceedings of Machine Learning Research},
  month = 	 {21--27 Jul},
  publisher =    {PMLR},
  url = 	 {https://proceedings.mlr.press/v235/cao24c.html}
}

@inproceedings{xie2025seeking,
  title={Seeking proxy point via stable feature space for noisy correspondence learning},
  author={Xie, Yucheng and Cai, Songyue and Tong, Tao and Hu, Ping and Zhu, Xiaofeng},
  booktitle={Proceedings of the Thirty-Fourth International Joint Conference on Artificial Intelligence},
  pages={2072--2080},
  year={2025}
}

@article{zhang2025prompts,
  title={Prompts Libra: Enhanced Image Outpainting Diffusion Model with Balanced Bimodal Guidance},
  author={Zhang, Zongyan and Chen, CL Philip and Su, Zepeng and Zhang, Tong},
  journal={IEEE Transactions on Circuits and Systems for Video Technology},
  year={2025},
  publisher={IEEE}
}

@article{CUI2026111991,
title = {MGCM: Multi-modal graph convolutional mamba for cancer survival prediction},
journal = {Pattern Recognition},
volume = {169},
pages = {111991},
year = {2026},
issn = {0031-3203},
doi = {https://doi.org/10.1016/j.patcog.2025.111991},
url = {https://www.sciencedirect.com/science/article/pii/S003132032500651X},
author = {Jiaqi Cui and Yilun Li and Dinggang Shen and Yan Wang}
}

@article{si2025unified,
  title={A unified framework of data augmentation using large language models for text-based cross-modal retrieval},
  author={Si, Lijia and Guo, Caili and Li, Zheng and Yang, Yang},
  journal={Pattern Recognition},
  pages={111755},
  year={2025},
  publisher={Elsevier}
}

@article{ke2025cross,
  title={Cross-modal independent matching network for image-text retrieval},
  author={Ke, Xiao and Chen, Baitao and Yang, Xiong and Cai, Yuhang and Liu, Hao and Guo, Wenzhong},
  journal={Pattern Recognition},
  volume={159},
  pages={111096},
  year={2025},
  publisher={Elsevier}
}

@article{li2025multi,
  title={Multi-view visual semantic embedding for cross-modal image--text retrieval},
  author={Li, Zheng and Guo, Caili and Wang, Xin and Zhang, Hao and Hu, Lin},
  journal={Pattern Recognition},
  volume={159},
  pages={111088},
  year={2025},
  publisher={Elsevier}
}

@article{dang2025disentangled,
  title={Disentangled noisy correspondence learning},
  author={Dang, Zhuohang and Luo, Minnan and Wang, Jihong and Jia, Chengyou and Han, Haochen and Wan, Herun and Dai, Guang and Chang, Xiaojun and Wang, Jingdong},
  journal={IEEE Transactions on Image Processing},
  year={2025},
  publisher={IEEE}
}

@inproceedings{dang2024noisy,
  title={Noisy correspondence learning with self-reinforcing errors mitigation},
  author={Dang, Zhuohang and Luo, Minnan and Jia, Chengyou and Dai, Guang and Chang, Xiaojun and Wang, Jingdong},
  booktitle={Proceedings of the AAAI Conference on Artificial Intelligence},
  volume={38},
  number={2},
  pages={1463--1471},
  year={2024}
}

@inproceedings{wang2025noisy,
  title={Noisy Correspondence Rectification via Asymmetric Similarity Learning},
  author={Wang, Yunbo and Wu, YuJie and Dai, Zhien and Tian, Can and Long, Jun and Chen, Jianhai},
  booktitle={Proceedings of the AAAI Conference on Artificial Intelligence},
  volume={39},
  number={20},
  pages={21384--21392},
  year={2025}
}

@article{zhang2019epigenetic,
  title={Epigenetic-mediated downregulation of zinc finger protein 671 (ZNF671) predicts poor prognosis in multiple solid tumors},
  author={Zhang, Jian and Zheng, Ziqi and Zheng, Jieling and Xie, Tao and Tian, Yunhong and Li, Rong and Wang, Baiyao and Lin, Jie and Xu, Anan and Huang, Xiaoting and others},
  journal={Frontiers in Oncology},
  volume={9},
  pages={342},
  year={2019},
  publisher={Frontiers Media SA}
}

@article{michaelidou2015clinical,
  title={Clinical relevance of the deregulated kallikrein-related peptidase 8 mRNA expression in breast cancer: a novel independent indicator of disease-free survival},
  author={Michaelidou, Kleita and Ardavanis, Alexandros and Scorilas, Andreas},
  journal={Breast cancer research and treatment},
  volume={152},
  pages={323--336},
  year={2015},
  publisher={Springer}
}

@article{arabkari2023mirna,
  title={miRNA-378 is downregulated by XBP1 and inhibits growth and migration of luminal breast cancer cells},
  author={Arabkari, Vahid and Barua, David and Hossain, Muhammad Mosaraf and Webber, Mark and Smith, Terry and Gupta, Ananya and Gupta, Sanjeev},
  journal={International Journal of Molecular Sciences},
  volume={25},
  number={1},
  pages={186},
  year={2023},
  publisher={MDPI}
}

@article{LAN2026113025,
title = {Contrastive adversarial tuning: Enhancing discriminability and robustness of LLMs for emotion recognition in conversation},
journal = {Pattern Recognition},
volume = {174},
pages = {113025},
year = {2026},
issn = {0031-3203},
doi = {https://doi.org/10.1016/j.patcog.2025.113025},
url = {https://www.sciencedirect.com/science/article/pii/S0031320325016887},
author = {Kankan Lan and C. L. Philip Chen and Zongyan Zhang and Tong Zhang}
}

\end{document}